\documentclass[11pt]{article}

\usepackage{booktabs}
\usepackage{colortbl}
\usepackage[table]{xcolor}
\definecolor{lightblue}{RGB}{220,235,250}
\definecolor{boxframe}{RGB}{175,210,240} 
\definecolor{boxback}{RGB}{242, 242, 245} 
\definecolor{systemcolor}{RGB}{0, 90, 160}    
\definecolor{usercolor}{RGB}{0, 140, 70}      
\definecolor{assistantcolor}{RGB}{130, 50, 200} 
\usepackage{amsmath}
\usepackage{amssymb}
\usepackage{subcaption}
\usepackage{algorithm}
\usepackage{algorithmic}
\usepackage{dsfont}
\usepackage{enumitem}
\usepackage[most]{tcolorbox} 
\newcommand{\mathbbm}[1]{\mathds{#1}}
\usepackage{marvosym}
\usepackage{footmisc}
\usepackage[final]{acl}

\usepackage{times}
\usepackage{latexsym}

\usepackage[T1]{fontenc}

\usepackage[utf8]{inputenc}

\usepackage{microtype}

\usepackage{inconsolata}

\usepackage{graphicx}

\title{Beyond Majority Voting: Towards Fine-grained and More Reliable\\ Reward Signal for Test-Time Reinforcement Learning}

\author{
    Weiqin Wang$^{1}$, 
    Yile Wang$^{1,}$\textsuperscript{\Letter}, 
    Kehao Chen$^{2}$, 
    Hui Huang$^{1}$ \\
    \textsuperscript{1}College of Computer Science and Software Engineering, Shenzhen University \\
    \textsuperscript{2}College of Computer and Data Science, Fuzhou University \\
    \texttt{here1swqw@gmail.com, wangyile@szu.edu.cn}
}

\begin{document}
\maketitle

\DefineFNsymbols*{1}{\Letter}
\setfnsymbol{1}
\renewcommand{\thefootnote}{\fnsymbol{footnote}} 
    \footnotetext[1]{Corresponding author.}
\renewcommand{\thefootnote}{\arabic{footnote}}

\begin{abstract}

Test-time reinforcement learning mitigates the reliance on annotated data by using \textit{majority voting} results as pseudo-labels, emerging as a complementary direction to reinforcement learning with verifiable rewards (RLVR) for improving reasoning ability of large language models (LLMs). However, this voting strategy often induces \textit{confirmation bias} and suffers from \textit{sparse rewards}, limiting the overall performance. In this work, we propose \underline{s}ubgroup-specific step-wise \underline{co}nfidence-weighted \underline{p}seudo-label \underline{e}stimation (\textsc{SCOPE}), a framework integrating model confidence and dynamic subgroup partitioning to address these issues. Specifically, \textsc{SCOPE} integrates the proposed step-wise confidence into pseudo-label estimation, prioritizing high-quality reasoning paths over simple frequency count. Furthermore, it dynamically partitions the candidate outputs pool into independent subgroups by balancing reasoning quality against exploration diversity. By deriving local consensus via repeat sampling for each subgroup, \textsc{SCOPE} provides diverse supervision targets to encourage broader exploration. We conduct experiments across various models and benchmarks, experimental results show that \textsc{SCOPE} consistently outperforms recent baselines. Notably, \textsc{SCOPE} achieves relative improvements of 13.1\% on challenging AIME 2025 and 8.1\% on AMC. The code is released at \url{https://github.com/szu-tera/SCOPE}.

\end{abstract}

\section{Introduction}

Reinforcement learning (RL) has become an important paradigm in improving the reasoning capability of large language models (LLMs). The paradigm of reinforcement learning with verifiable rewards (RLVR) has also been used in seminal models such as DeepSeek-R1~\cite{guo2025deepseek}, Qwen3~\cite{yang2025qwen3}, and OpenAI's o1~\cite{jaech2024openai}. From the perspective of training data, RLVR is similar to supervised fine-tuning that requires ground-truth labels to guide the iterative policy learning process and thus elicit the strong reasoning ability of LLMs~\citep{wen2025reinforcement,su2025crossing,tang2025towards}.


\begin{figure}[t!]
    \centering
    \includegraphics[width=\linewidth]{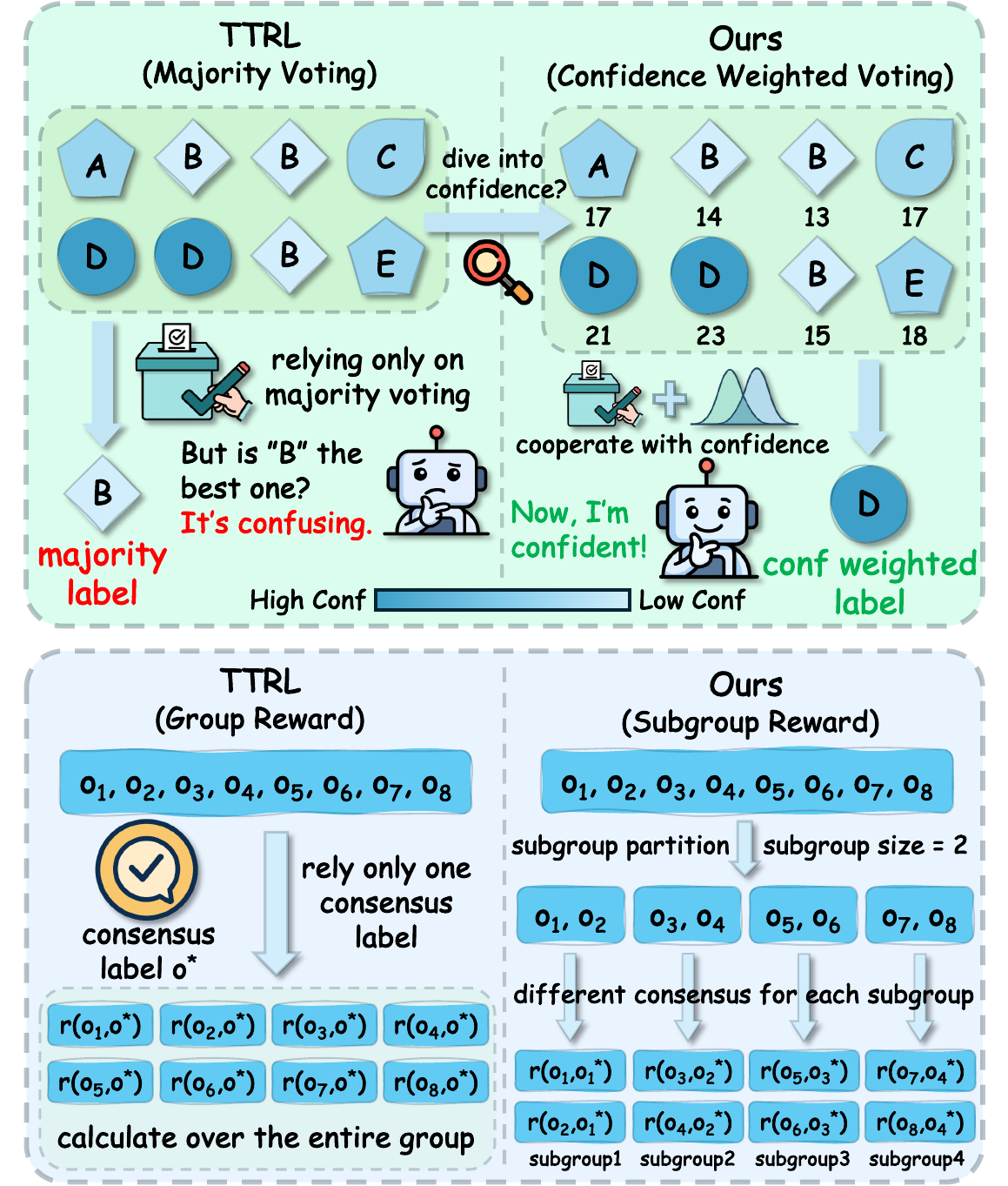}
    \caption{Illustration of the difference betweens \textsc{TTRL}~\cite{zuo2025ttrl} and our method. Top: consensus label estimation with step-wise confidence weighting. Bottom: group partition and reward calculation using subgroup-specific consensus labels.}
    \label{fig:intro}
\end{figure}

However, dependence on extensive manual labeling is costly and inefficient, especially for large-scale or complex tasks. Thus, there are studies that try to enable reinforcement learning of LLMs without supervision. \citet{zuo2025ttrl} proposed test-time reinforcement learning (\textsc{TTRL}) that samples multiple responses from the policy model and then employs majority voting to obtain a consensus label to replace a predefined ground-truth label, enabling LLMs to be trained directly in real-world settings without supervision.

While \textsc{TTRL} provides a straightforward and effective framework for unsupervised RL, its reliance on majority voting for pseudo-label generation represents a significant bottleneck. This coarse-grained estimation process treats all votes equally regardless of their underlying confidence, leading to two critical issues: (1) \textit{confirmation bias}, where the model risks reinforcing its own errors from incorrect label estimation~\citep{pseudoLabel2019,Prabhu_2021_ICCV,wang-etal-2025-ranked}, particularly when the majority consensus aligns with a plausible but incorrect answer; and (2) \textit{sparse rewards}, as the binary nature of voting fails to capture dense, fine-grained signals essential for fine-tuning~\citep{lightman2024lets}. While recent advances have successfully leveraged step-wise mechanisms to optimize test-time reasoning efficiency~\cite{huang2026sat}, utilizing such fine-grained signals to improve test-time reward reliability remains largely underexplored.

To mitigate the above limitations of \textsc{TTRL}, we propose \underline{s}ubgroup-specific step-wise \underline{co}nfidence-weighted \underline{p}seudo-label \underline{e}stimation (\textsc{SCOPE}). First, \textsc{SCOPE} introduces step-wise confidence for pseudo-label estimation to address the issue of potentially incorrect label estimation. Figure~\ref{fig:intro} (Top) shows an example of eight responses where the vote counts for solution A$\sim$E are 1, 3, 1, 2, and 1, respectively. The majority voting strategy selects solution B, as it receives the highest number of votes. However, a step-wise confidence analysis reveals that solution B exhibit low confidence (i.e., high uncertainty). In contrast, solution D demonstrates higher accumulated confidence and is the correct solution, though having less vote counts. Second, we design a subgroup-specific pseudo-label estimation strategy to alleviate the issue of sparse rewards. Unlike \textsc{TTRL}, which assigns a uniform pseudo-label to all sampled responses for reward calculation, our method partitions the responses into distinct subgroups, assigning a separate label to each subgroup, as shown in Figure~\ref{fig:intro} (Bottom). Moreover, we employ Pareto optimization~\cite{pareto1964cours} to automatically select the optimal subgroup size during training.

We validate our method on advanced LLMs, including LLaMA3.1~\citep{grattafiori2024llama}, Qwen2.5~\citep{yang2024qwen2}, and Qwen3~\citep{yang2025qwen3} with different parameter sizes. Empirical results demonstrate that \textsc{SCOPE} consistently outperforms baselines. In particular, applying \textsc{SCOPE} to Qwen3-8B yields an improvement on AIME 2024~\citep{li2024numinamath} of 10.48\%, with an average gain of 6.85\% across all benchmarks.

Our contributions are summarized as follows:

\begin{itemize}[itemsep=3pt, topsep=2pt, parsep=0pt, partopsep=0pt]
    \item We propose \textsc{SCOPE}, a novel test-time reinforcement learning framework that leverages step-wise confidence weighting and subgroup-specific label estimation to mitigate the limitations of sparse rewards and confirmation bias in unsupervised RL.
    
    \item We introduce step-wise confidence to recover correct answers in the minority and Pareto-optimized subgroups to balance reward density and estimation accuracy during training.
    
    \item Extensive experiments demonstrate the effectiveness of \textsc{SCOPE} in improving reasoning capabilities, achieving superior performance across multiple benchmarks.
\end{itemize}

\begin{figure*}[t!]
  \centering
  \includegraphics[width=1\textwidth]{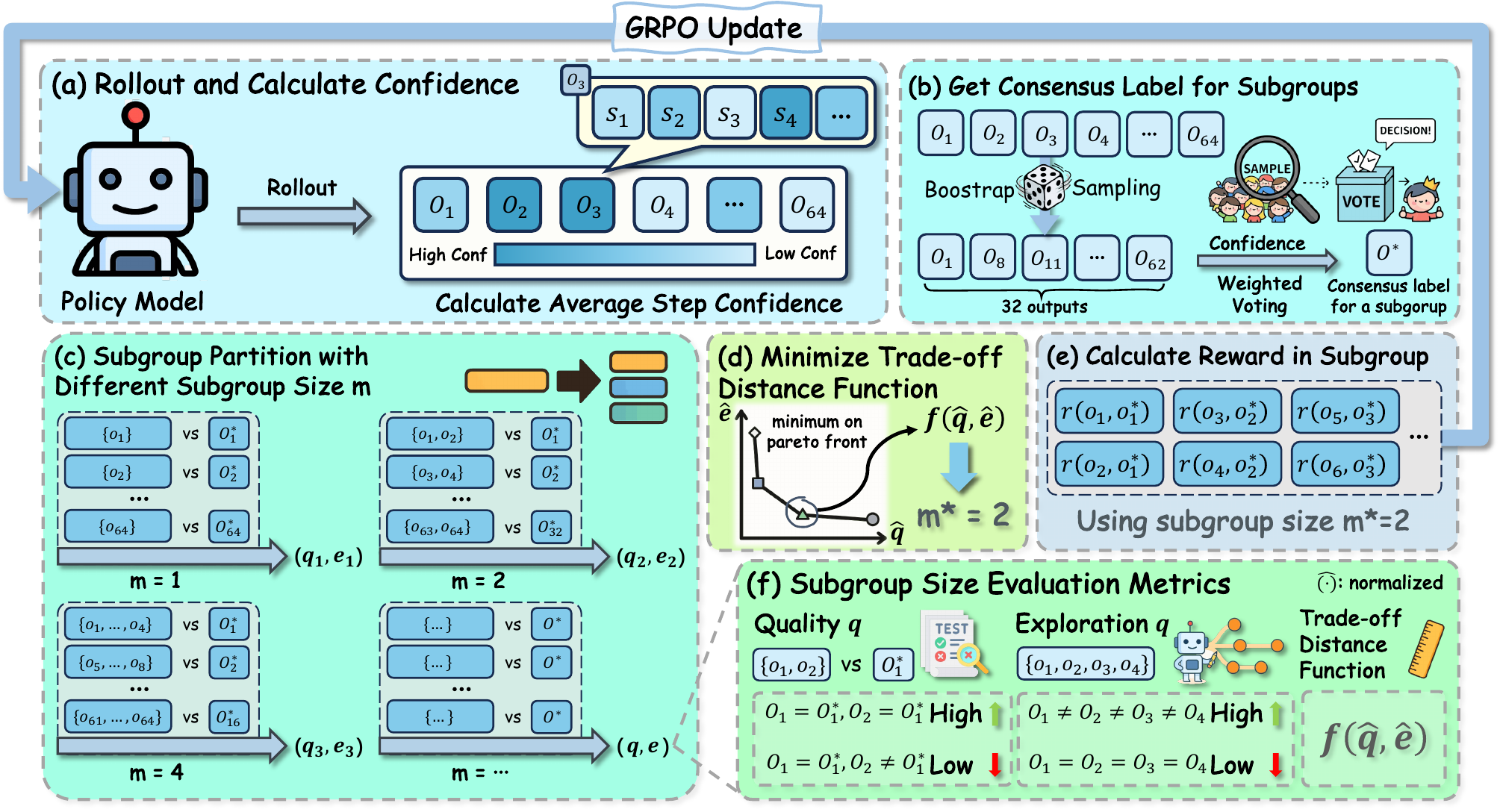}
  \caption{Overview of the \textsc{SCOPE} framework. The process involves \textbf{(a)} generating responses with step-wise confidence, \textbf{(b)} estimating consensus labels via weighted voting, \textbf{(c)} evaluating different subgroup partitions, \textbf{(d)} employing Pareto optimization  to select the optimal subgroup size $m^*$ by balancing \textbf{(f)} quality and exploration metrics, and \textbf{(e)} computing rewards using the optimized subgroup strategy for model updates.}
  \label{fig:method}
\end{figure*}

\section{Preliminaries}

\subsection{Group Relative Policy Optimization}

Group Relative Policy Optimization (GRPO; \citealp{shao2024deepseekmath}) estimates the advantage of a policy by leveraging group-wise relative rewards, eliminating the need for a separate value function. Given a group of candidate outputs \( \{o_i\}_{i=1}^{\left| \mathcal{G} \right|} \) sampled from the old policy \(\pi_{\theta_{\text{old}}}\) for an input \( \mathbf{x} \), GRPO computes the advantage \( \mathcal{A}_i \) for each response \( o_i \):
\begin{equation}
    \mathcal{A}_i = \frac{r(o_i) - \mu_g}{\sigma_g + \epsilon},
\end{equation}
and \( r(o_i) \) denotes the reward for response \( o_i \). \( \mu_g, \sigma_g \) represent the mean and standard deviation of the rewards within the group, respectively. \( \epsilon \) is a small constant added for numerical stability. Subsequently, the policy model \(\pi_\theta\) is optimized by maximizing the following surrogate objective:

\begin{equation}
\scalebox{0.85}{$
    \begin{aligned}
        \mathcal{J}_{\text{GRPO}}(\theta) &= \mathbb{E}_{\mathbf{x}, \{o_i\} \sim \pi_{\text{old}}} \Bigg[ \frac{1}{\left| \mathcal{G} \right|} \sum_{i=1}^{\left| \mathcal{G} \right|} \frac{1}{|o_i|} \sum_{t=1}^{|o_i|} \Big( \\ 
        &\quad \min \Big[ \rho_{i,t} \mathcal{A}_i, \text{clip}(\rho_{i,t}, 1-\epsilon, 1+\epsilon) \mathcal{A}_i \Big] \\ 
        &\quad - \beta \mathbb{D}_{\text{KL}}[\pi_\theta || \pi_{\text{ref}}]_t \Big) \Bigg], 
    \end{aligned}$}
    \label{eq:grpo_obj}
\end{equation}
where \( \rho_{i,t} = \frac{\pi_\theta(o_{i,t}|\mathbf{x}, o_{i,<t})}{\pi_{\theta_{\text{old}}}(o_{i,t}|\mathbf{x}, o_{i,<t})} \) represents the probability ratio between the current and old policies at step \( t \), \( \beta \) is the coefficient for the KL divergence penalty, and \( \pi_{\text{ref}} \) is the reference model used to prevent excessive policy deviation.

\subsection{Test-Time Reinforcement Learning}

\citet{zuo2025ttrl} proposed TTRL to mitigate the dependence on ground-truth labels \( g \), which are traditionally required in the reward function \( r(o_i, g) \). The core idea is to replace external supervision by leveraging the majority voting result over a group of sampled outputs \( \{o_i\}_{i=1}^{\left| \mathcal{G} \right|} \) to derive a consensus label \( o^* \):
\begin{equation}
    o^* = \operatorname*{argmax}_o \sum_{i=1}^{\left| \mathcal{G} \right|} \mathbbm{1}(o_i = o),
\end{equation}
where consensus output \( o^* \) serves as a \textit{pseudo-label}, allowing the reward to be computed through \( r(o_i, o^*) \) without requiring ground-truth labels.

\subsection{Token Confidence}
Recent studies indicate that LLMs exhibit varying levels of confidence during the reasoning process, and token confidence is often utilized as a metric to quantify the local certainty of a language model's prediction at a specific step~\cite{fu2025deepParadigm}. Given the predicted probability distribution at position \( i \), the token confidence \( \mathcal{C}_i \) is defined as the negative average log-probability of the top-\( k \) most probable tokens:
\begin{equation}
    \mathcal{C}_i = -\frac{1}{k} \sum_{j=1}^{k} \log P_i(j),
\label{eq:pij}
\end{equation}
where \( P_i(j) \) denotes the probability of the \( j \)-th candidate among the top-\( k \) tokens during decoding. Conceptually, a higher \( \mathcal{C}_i \) implies that the probability mass is concentrated on a few tokens with a peaked distribution, indicating high certainty. Conversely, a lower value reflects a flatter distribution, suggesting that the model is uncertain about the next token prediction.

\section{Method}

Figure~\ref{fig:method} shows the overall pipeline of \textsc{SCOPE}. We begin by introducing the \textit{average step confidence} (\S\ref{subsec:step-conf}) and its calculation method. Next, we propose the concept of \textit{subgroup} (\S\ref{subsec:sub-group}) and its detailed formulation. Subsequently, we describe how to adaptively determine the optimal subgroup size during training via \textit{Pareto optimization} (\S\ref{subsec:adaptive}). Finally, we provide the \textit{unified algorithm} of \textsc{SCOPE} that integrates the above key components (\S\ref{subsec:overview}).

\subsection{Thinking with Step Confidence}
\label{subsec:step-conf}

We first propose \textbf{Average Step Confidence} to capture fine-grained uncertainty of LLMs during reasoning. This metric is designed to balance the reward signals with different granularity, avoiding the excessive noise often observed in raw token-level probabilities while retaining structural precision. Specifically, we decompose a response \( o_i \) into a sequence of reasoning steps by newline delimiter ``\textbackslash n\textbackslash n'', denote as  $o_i = {s_1, ..., s_k, ..., s_{\left| \mathcal{L} \right|}}$, with length $\left| \mathcal{L} \right|$. The confidence for each step \( s_k \) is then calculated by averaging the confidence scores of its constituent tokens:

\begin{equation}
    \mathcal{C}_{s_k} = \frac{1}{N_k} \sum_{t=1}^{N_k} \mathcal{C}_t,
\end{equation}
where \( N_k \) denotes the number of tokens in the \( k \)-th step \( s_k \), and \( \mathcal{C}_t \) represents the confidence score of the \( t \)-th token within that step, as calculated in Eq.~\ref{eq:pij}. To quantify the overall certainty of the response \( o_i \), we compute the average step confidence by aggregating the step-level scores: 
\begin{equation}
    \mathcal{C}_{\text{AvgStep}}^{(i)} = \frac{1}{\left| \mathcal{L} \right|} \sum_{k=1}^{\left| \mathcal{L} \right|} \mathcal{C}_{s_k},
\end{equation}
where \( \left| \mathcal{L} \right| \) is the total number of steps in \( o_i \). Finally, as shown in Figure~\ref{fig:method}(b), we employ these confidence scores as weights to estimate the consensus label. Unlike naive majority voting, our method assigns higher importance to responses with higher average step confidence:
\begin{equation}
    o^* = \operatorname*{argmax}_{y} \sum_{i=1}^{\left| \mathcal{G} \right|} \mathcal{C}_{\text{AvgStep}}^{(i)} \cdot \mathbbm{1}\left[\text{Ans}(o_i) = y\right],
\end{equation}
where \( \text{Ans}(o_i) \) denotes the final answer extracted from response \( o_i \), and the maximization is performed over all unique candidate answers \( y \). Here, \( o^* \) represents the consensus label derived via confidence-weighted voting. By prioritizing candidates based on reasoning certainty rather than simple frequency, \( o^* \) serves as a robust, confidence-aware target for reward calculation, superseding majority voting. Besides step-level confidence, we compare different confidence granularities in \S\ref{subsec:conf_granularity}.

\subsection{Fine-grained Rewards within Subgroups}
\label{subsec:sub-group}

Instead of relying on a single consensus derived from the entire set of outputs \( \{ o_i \}_{i=1}^{\left| \mathcal{G} \right|} \), which can be sparse, we define a granular unit for reward computation, referred to as a \textbf{Subgroup}. As illustrated in Figure~\ref{fig:method}(c), we partition the global pool of \( \left| \mathcal{G} \right| \) generated responses into distinct subgroups, each containing \( m = \left| \mathcal{G} \right|/n \) outputs. The set of subgroups can be formally denoted as: 
\begin{equation}
    \mathcal{S} = \Big\{ S_j = \{ o_{(j-1)m + 1}, \ldots, o_{jm} \} \Big\}_{j=1}^{n}.
\end{equation}

To derive the consensus label \( o^*_j \) for each subgroup \( S_j \), we employ an independent estimation strategy. Specifically, as depicted in Figure~\ref{fig:method}(b), for each of the \( n \) subgroups, we perform bootstrap sampling from the global pool of generated responses \( \{o_i\}_{i=1}^{\left| \mathcal{G} \right|} \) to construct a candidate set. We then apply the confidence-weighted voting mechanism described in \S\ref{subsec:step-conf} to this set to determine \( o^*_j \). Consequently, the label estimation process is executed \( n \) times, each corresponding to a distinct subgroup. This strategy enables subgroups to explore diverse reasoning paths while ensuring each receives a robust consensus target derived from the global distribution.

Finally, as shown in Figure~\ref{fig:method}(e), the correctness reward for each output \( o \in S_j \) is computed against its corresponding subgroup-specific consensus \( o^*_j \):
\begin{equation}
    r(o, o^*_j) = \mathbbm{1}\left[\text{Ans}(o) = o^*_j\right].
\end{equation}

\subsection{Automatic Subgroup Size Selection with Pareto Optimization}
\label{subsec:adaptive}

Having defined the subgroup mechanism, we now detail the procedure for automatically selecting the optimal subgroup size \( m \) during training. As illustrated in Figure~\ref{fig:method}(f), this selection process aims to balance two competing objectives: (1) \textbf{Reasoning Quality}: ensuring output correctness by maximizing alignment with the local consensus; and (2) \textbf{Exploration}: preserving solution diversity to avoid overconfidence and mode collapse.

To quantify these objectives, we formulate two metrics. The first is the \textit{quality rate} \( q \), which measures the consistency of outputs with their subgroup consensus:
\begin{equation}
    q = \frac{1}{\left| \mathcal{G} \right|} \sum_{j=1}^{n} \sum_{l=1}^{m} \mathbbm{1}\!\left[\text{Ans}(o_{(j-1)m + l}) = o_j^*\right],
\end{equation}
where \( \left| \mathcal{G} \right| \) is the total number of generated outputs, \( n \) is the number of subgroups, and \( o_j^* \) denotes the consensus label of the \( j \)-th subgroup. The second metric is the \textit{exploration rate} \( e \), defined as the proportion of unique consensus labels discovered:
\begin{equation}
    e = \frac{\bigl|\{\, o_1^*, o_2^*, \dots, o_n^* \,\}\bigr|}{n},
\end{equation}
where \( o_j^* \) is the consensus label of the \( j \)-th subgroup, and \( |\cdot| \) denotes the cardinality of the set of unique consensus outputs across all subgroups.

As depicted in Figure~\ref{fig:method}(c), we evaluate a set of candidate subgroup sizes (e.g., \( m=1, 2, 4, \dots \)) by computing their corresponding pairs \( (q_k, e_k) \). Inspired by previous work \citep{NEURIPS2024_89f39d0b,lou2025adacot}, we formulate this selection process as an optimization problem. In particular, we construct a Pareto front \(\{(q_k, e_k)\}_{k=1}^{P}\) from these candidates to identify non-dominated solutions. To select the final optimal size \( m^* \), we compute a \textit{trade-off distance} for each Pareto-optimal point, as shown in Figure~\ref{fig:method}(d). We first normalize the metrics:
\begin{equation}
    \hat{q}_k = \frac{q_k - q_{\min}}{q_{\max} - q_{\min}},
\end{equation}
\begin{equation}
    \hat{e}_k = \frac{e_k - e_{\min}}{e_{\max} - e_{\min}}.
\end{equation}

Then, we compute the weighted trade-off distance \( d_k \) for each candidate point to the ideal state:
\begin{equation}
    d_k = \sqrt{\lambda (1 - \hat{q}_k)^2 + (1 - \lambda)(1 - \hat{e}_k)^2 },
\label{eq:lambda}
\end{equation}
where \( \lambda \in [0, 1] \) is a trade-off parameter that controls the preference between quality and exploration. We set \( \lambda = 0.7 \) in our experiments, and a detailed analysis of this parameter selection is provided in \S\ref{subsec:lambda_sensitivity}. Finally, we select the optimal subgroup size \( m^* \) that minimizes this distance:
\begin{equation}
    m^* = \operatorname*{argmin}_{m_k} d_k.
\end{equation}

\subsection{The Unified Algorithm}
\label{subsec:overview}

We summarize the complete training procedure of \textsc{SCOPE} in Algorithm~\ref{alg:scope}. In each iteration, the policy model first generates a pool of candidate responses for a given input, and we compute the average step confidence for each response (\S\ref{subsec:step-conf}). To dynamically balance the trade-off between reasoning quality and exploration, \textsc{SCOPE} evaluates multiple subgroup configurations and selects the optimal subgroup size \( m^* \) via Pareto optimization (\S\ref{subsec:adaptive}). Subsequently, the responses are partitioned into subgroups based on \( m^* \), where local consensus labels are derived using step-wise confidence-weighted voting with bootstrap sampling (\S\ref{subsec:sub-group}). Finally, fine-grained rewards are computed against local targets to update the policy via GRPO.

\begin{algorithm}[t]
    \caption{Training Iteration of \textsc{SCOPE}}
    \label{alg:scope}
    \footnotesize  
    \renewcommand{\baselinestretch}{1.25}
    \selectfont
    
    \begin{algorithmic}[1]
        \REQUIRE Input dataset \(\mathcal{D}\), Policy \(\pi_\theta\), Candidate sizes \(\mathcal{M}\).
        
        \STATE Sample \(\mathbf{x} \sim \mathcal{D}\) and rollout \(\{o_i\}_{\smash{i=1}^{|\mathcal{G}|}} \sim \pi_\theta(\cdot|\mathbf{x})\).
        \STATE Calculate average step confidence \(\mathcal{C}_{\smash{\text{AvgStep}}}^{(i)}\) for each output.
        
        \vspace{2pt}
        \STATE \textbf{// Automatic Subgroup Size Selection}
        \STATE Evaluate quality \(q\) and exploration \(e\) for all \(m \in \mathcal{M}\).
        \STATE Select optimal subgroup size \(m^*\) by minimizing the trade-off distance on the Pareto front.
        
        \vspace{2pt}
        \STATE \textbf{// Subgroup-specific Reward Computation}
        \STATE Partition \(\{o_i\}_{i=1}^{|\mathcal{G}|}\) into subgroups \(\{S_j\}\) of size \(m^*\).
        \FOR{each subgroup \(S_j\)}
            \STATE Derive local consensus \(o_j^*\) via bootstrap sampling and confidence-weighted voting.
            \STATE Set rewards \(r_i \leftarrow \mathbbm{1}[\text{Ans}(o_i) = \text{Ans}(o_j^*)]\) for \(o_i \in S_j\).
        \ENDFOR
        
        \vspace{2pt}
        \STATE Update \(\pi_\theta\) using GRPO objective with computed rewards.
    \end{algorithmic}
\end{algorithm}

\setlength{\tabcolsep}{7pt}  
\begin{table*}[t!]
  \centering
  \small
  \scalebox{0.87}{
    \begin{tabular}{lccccc}
      \toprule
      \textbf{Models} & \textbf{AIME 2024} & \textbf{AIME 2025} & \textbf{AMC} & \textbf{MATH-500}  & \textbf{Avg} \\
      \midrule
      \rowcolor{gray!20}
      \multicolumn{6}{c}{\textsc{I. Lightweight-sized Models}} \\
      \midrule
      \rowcolor{gray!20}
      \multicolumn{6}{c}{\textsc{Qwen2.5-Math-1.5B}} \\
      \textsc{Qwen2.5-Math-1.5B}~\cite{yang2024qwen2}                          & \phantom{0}7.92$_{\pm0.11}$    & \phantom{0}3.12$_{\pm0.08}$    & 26.58$_{\pm0.29}$ & 32.21$_{\pm0.33}$    & 17.46    \\
      \textit{w/} \textsc{INTUITOR}~\cite{zhao2025learning}                  & \phantom{0}6.88$_{\pm0.12}$   & \phantom{0}5.50$_{\pm0.07}$    & 39.08$_{\pm0.25}$ & 64.26$_{\pm0.21}$    & 28.93   \\
      \textit{w/} \textsc{TTRL}~\cite{zuo2025ttrl}                   & \underline{16.48}$_{\pm0.12}$   & \phantom{0}\underline{9.86}$_{\pm0.12}$    & \underline{48.87}$_{\pm0.17}$ & \underline{72.58}$_{\pm0.16}$    & \underline{36.95}    \\
      \rowcolor{lightblue!100}
      \textit{w/} \textsc{SCOPE} (Ours)                   & \textbf{22.50$_{\pm0.11}$}    & \textbf{14.90$_{\pm0.03}$}    & \textbf{51.20$_{\pm0.15}$}    & \textbf{76.85$_{\pm0.09}$}    & \textbf{41.36}    \\
      \rowcolor{lightblue!100}
      $\Delta$                                    & $\uparrow6.02/36.5\%$     & $\uparrow5.04/51.1\%$     & $\uparrow2.33/4.8\%$\phantom{$0$}    & $\uparrow4.27/5.9\%$    & $\uparrow4.40/11.9\%$    \\
      \midrule
      
      \rowcolor{gray!20}
      \multicolumn{6}{c}{\textsc{Qwen3-1.7B}} \\
      \textsc{Qwen3-1.7B}~\cite{yang2025qwen3}                                    & 11.45$_{\pm0.11}$     & 11.53$_{\pm0.11}$    & 39.30$_{\pm0.26}$    & 73.23$_{\pm0.17}$    & 33.88    \\
      \textit{w/} \textsc{INTUITOR}~\cite{zhao2025learning}                  & 15.18$_{\pm0.04}$   & 11.42$_{\pm0.08}$    & 45.58$_{\pm0.19}$ & 76.11$_{\pm0.13}$    & 37.07   \\
      \textit{w/} \textsc{Co-rewarding-I}~\cite{zhang2025co}                          & 16.42$_{\pm0.08}$     & 12.19$_{\pm0.10}$     & \underline{51.99}$_{\pm0.13}$    & \underline{78.91}$_{\pm0.09}$    & 39.88   \\ 
      \textit{w/} \textsc{TTRL}~\cite{zuo2025ttrl}                            & \underline{19.37}$_{\pm0.09}$     & \underline{19.23}$_{\pm0.01}$   & 50.45$_{\pm0.08}$    & 78.18$_{\pm0.07}$    & \underline{41.91}    \\
      \rowcolor{lightblue!100}
      \textit{w/} \textsc{SCOPE} (Ours)                            & \textbf{21.66$_{\pm0.05}$}     & \textbf{19.71$_{\pm0.02}$}     & \textbf{53.46$_{\pm0.05}$}    & \textbf{81.27$_{\pm0.05}$}    & \textbf{44.02}    \\
      \rowcolor{lightblue!100}
      $\Delta$                                    & $\uparrow2.29/11.8\%$     & $\uparrow0.48/2.5\%$\phantom{$0$}     & $\uparrow3.01/6.0\%$\phantom{$0$}    & $\uparrow3.09/4.0\%$\phantom{$0$}    & $\uparrow2.11/5.0\%$\phantom{$0$}    \\
      \midrule

      \midrule
      \rowcolor{gray!20}
      \multicolumn{6}{c}{\textsc{II. Medium-sized Models}} \\
      \midrule
      \rowcolor{gray!20}
      \multicolumn{6}{c}{\textsc{LLaMA3.1-8B-Instruct}} \\
      \textsc{LLaMA3.1-8B-Inst}~\cite{grattafiori2024llama}                                 & \phantom{0}6.46$_{\pm0.08}$     & \phantom{0}0.00$_{\pm0.00}$     & 19.27$_{\pm0.20}$    & 49.06$_{\pm0.26}$    & 18.70    \\
      \textit{w/} \textsc{RLPR}~\cite{yu2025rlpr}                  & \underline{10.00}$_{\pm0.11}$   & \phantom{0}\underline{0.96}$_{\pm0.03}$    & 24.92$_{\pm0.21}$ & 54.70$_{\pm0.22}$    & 22.64   \\
      \textit{w/} \textsc{TTRL}~\cite{zuo2025ttrl}                         & \phantom{0}9.56$_{\pm0.03}$     & \phantom{0}\underline{0.96}$_{\pm0.02}$     & \underline{32.08}$_{\pm0.12}$    & \textbf{62.93$_{\pm0.10}$}    & \underline{26.38}    \\
      \rowcolor{lightblue!100}
      \textit{w/} \textsc{SCOPE} (Ours)                         & \textbf{14.37$_{\pm0.02}$}     & \textbf{\phantom{0}1.44$_{\pm0.04}$}     & \textbf{35.24$_{\pm0.10}$}    & \underline{61.67}$_{\pm0.13}$   & \textbf{28.18}    \\
      \rowcolor{lightblue!100}
      $\Delta$                                    & $\uparrow4.81/50.3\%$     & $\uparrow0.48/50.0\%$     & $\uparrow3.16/9.9\%$\phantom{$0$}    & $\downarrow1.26/2.0\%$\phantom{$0$}    & $\uparrow1.80/6.8\%$\phantom{$0$}    \\
      \midrule
      
      \rowcolor{gray!20}
      \multicolumn{6}{c}{\textsc{Qwen3-8B}} \\
      \textsc{Qwen3-8B}~\cite{yang2025qwen3}                                    & 26.45$_{\pm0.15}$     & 20.67$_{\pm0.09}$     & 59.50$_{\pm0.19}$    & 83.66$_{\pm0.12}$    & 47.57    \\
      \textit{w/} \textsc{EVOL-RL}~\cite{zhou2025evolving}                            & 41.22$_{\pm0.17}$     & \underline{30.34}$_{\pm0.09}$     & 69.62$_{\pm0.12}$    & \textbf{91.70$_{\pm0.04}$}    & \underline{58.22}   \\
      \textit{w/} \textsc{Co-rewarding-I}~\cite{zhang2025co}                          & 28.39$_{\pm0.03}$     & 21.74$_{\pm0.02}$     & \underline{71.39}$_{\pm0.03}$    & 88.34$_{\pm0.01}$    & 52.47   \\
      \textit{w/} \textsc{INTUITOR}~\cite{zhao2025learning}                  & 27.15$_{\pm0.13}$   & 26.19$_{\pm0.10}$    & 65.66$_{\pm0.15}$ & 89.20$_{\pm0.06}$    & 52.05   \\
      \textit{w/} \textsc{TTRL}~\cite{zuo2025ttrl}                            & \underline{47.13}$_{\pm0.05}$     & 27.40$_{\pm0.04}$     & 68.55$_{\pm0.10}$    & 89.74$_{\pm0.05}$    & 58.21    \\
      \rowcolor{lightblue!100}
      \textit{w/} \textsc{SCOPE} (Ours)                            & \textbf{52.70$_{\pm0.02}$}     & \textbf{31.00$_{\pm0.01}$}     & \textbf{74.09$_{\pm0.03}$}    & \underline{91.01}$_{\pm0.03}$    & \textbf{62.20}    \\
      \rowcolor{lightblue!100}
      $\Delta$                                    & $\uparrow5.57/11.8\%$    & $\uparrow3.60/13.1\%$     & $\uparrow5.54/8.1\%$\phantom{$0$}    & $\uparrow1.27/1.4\%$\phantom{$0$}    & $\uparrow3.99/6.9\%$\phantom{$0$}    \\
      \bottomrule
    \end{tabular}
  }
  \caption{Comparison between the baselines and our method. In each column, the best results are \textbf{in bold}, and the second-best results are \underline{underlined}. The $\Delta$ row reports the performance difference relative to the TTRL baseline.
}
  \label{tab:main_exp}
\end{table*}

\section{Experiments}

\subsection{Experimental Settings}

\paragraph{Models.} 
We evaluate the generalization ability of \textsc{SCOPE} through comprehensive experiments on a diverse set of widely used LLMs, covering a broad spectrum from lightweight to medium-scale parameter sizes. The models included in our experiments are as follows: \textsc{Qwen2.5-MATH-1.5B}~\citep{yang2024qwen2}, \textsc{Qwen3-1.7B}, \textsc{Qwen3-8B}~\citep{yang2025qwen3}, and \textsc{LLaMA3.1-8B-Instruct}~\citep{grattafiori2024llama}. The checkpoints are listed in Appendix~\ref{app:model_links}.

\paragraph{Evaluation.} 
We evaluate \textsc{SCOPE} on four representative benchmarks: \textbf{AIME 2024}~\citep{li2024numinamath}, \textbf{AIME 2025}~\citep{li2024numinamath}, \textbf{AMC}~\citep{li2024numinamath}, and \textbf{MATH-500} \citep{hendrycks2021measuring}. 
We list dataset examples in Appendix~\ref{app:data_examples} and evaluation details in Appendix~\ref{app:exp}.

\paragraph{Baselines.}

We adopt \textsc{\textbf{TTRL}}~\cite{zuo2025ttrl} with majority voting to get a consensus label as our primary baseline. Besides, we also compare following methods: (1) \textsc{\textbf{INTUITOR}}~\cite{zhao2025learning} uses self-certainty (internal confidence) as intrinsic rewards for unsupervised RL training; (2) \textsc{\textbf{RLPR}}~\citep{yu2025rlpr} uses the model's intrinsic probability of generating the reference answer as a reward signal to extend RLVR to general domains without external verifiers; (3) \textsc{\textbf{Co-rewarding-I}}~\cite{zhang2025co} generates reward signals by enforcing contrastive agreement between the model's reasoning outputs on original questions and their semantically equivalent rephrased counterparts; and (4) \textsc{\textbf{EVOL-RL}}~\citep{zhou2025evolving} combines majority voting for stability with a semantic novelty reward to encourage diverse reasoning paths. The training details are shown in Appendix~\ref{app:exp}.

\subsection{Main Results}

The results are shown in Table~\ref{tab:main_exp}. Overall, our method consistently achieves superior performance across all evaluated settings. We detail the analysis based on model scales below.

\paragraph{Lightweight-sized Models.} As shown in the first section of Table~\ref{tab:main_exp}, despite the limited reasoning capacity of Qwen2.5-Math-1.5B and Qwen3-1.7B, \textsc{SCOPE} yields substantial performance gains over \textsc{TTRL}. For Qwen2.5-Math-1.5B, \textsc{SCOPE} achieves an average score of 41.36, surpassing \textsc{TTRL} (36.95) by an absolute margin of 4.41 (+11.9\%). Notably, on the challenging AIME 2024 benchmark, \textsc{SCOPE} boosts the performance from 16.48 to 22.50, a remarkable relative improvement of 36.5\%. Similarly, for Qwen3-1.7B, \textsc{SCOPE} reaches an improved average score of 44.02, outperforming \textsc{TTRL} across all four benchmarks. These results confirm that even for smaller models which typically struggle with self-verification, our subgroup-specific confidence weighting effectively filters out incorrect reasoning paths, enabling lightweight models to perform significantly beyond their parameter scale.

\paragraph{Medium-sized Models.} 
The second section of Table~\ref{tab:main_exp} shows the results for medium-sized models, including LLaMA3.1-8B-Instruct and Qwen3-8B. On these stronger models, \textsc{SCOPE} also demonstrates remarkable performance. For LLaMA3.1-8B-Instruct, \textsc{SCOPE} achieves its largest relative gain on AIME 2024, improving performance from 9.56 of \textsc{TTRL} to 14.37, corresponding to a 50.3\% improvement. Despite a minor regression on the easier MATH-500, \textsc{SCOPE}’s substantial gains on competition-level datasets suggest it prioritizes complex reasoning over simpler tasks.On the strongest Qwen3-8B, \textsc{SCOPE} achieves absolute gains of 5.57 and 3.60 over \textsc{TTRL} on AIME 2024 and 2025, respectively. While maintaining comparable performance to \textsc{EVOL-RL} on the saturated MATH-500 benchmark, \textsc{SCOPE} establishes a dominant lead on competition-level tasks, surpassing \textsc{EVOL-RL} by 11.5\% on AIME 2024. The overall averaged performance across all benchmarks reaches 62.20, exhibiting a 6.9\% relative improvement over \textsc{TTRL}. These results suggest that stronger base models enable \textsc{SCOPE} to better leverage dense reward signals to rectify subtle errors in complex problem-solving scenarios.

\subsection{Ablation Study}
To demonstrate the effectiveness of our framework, we adopt several training settings: (1) \textbf{\textit{w/o Conf}}, which employs naive majority voting for pseudo-label estimation;  (2) \textbf{\textit{w/o Subgroup}}, which eliminates the subgroup partitioning strategy and computes the reward based on a single global consensus derived from the entire set of sampled outputs.

\setlength{\tabcolsep}{8pt}  
\begin{table}[t!]
  \centering
  \small
  \resizebox{\linewidth}{!}{
    \begin{tabular}{lcccc}
      \toprule
      \textbf{Models} & \textbf{AIME 2024}  & \textbf{AIME 2025} \\
      \rowcolor{gray!20}
      \midrule
      \multicolumn{3}{c}{\textsc{Qwen2.5-Math-1.5B}} \\
      \textsc{Qwen2.5-Math-1.5B}      & \phantom{0}7.92\phantom{ (-0.00)} & \phantom{0}3.12\phantom{ (-0.00)} \\
      \textit{w/ }\textsc{TTRL}       & 16.48\phantom{ (-0.00)} & \phantom{0}9.86\phantom{ (-0.00)} \\
      \rowcolor{lightblue!100}
      \textit{w/ }\textsc{SCOPE}                   & 22.50\phantom{ (-0.00)} & 14.90\phantom{ (-0.00)} \\
      - \textit{w/o Conf}                & 20.41 (-2.09) & 11.77 (-3.13) \\
      - \textit{w/o Subgroup}           & 16.67 (-5.83) & 11.05 (-3.85) \\
      \midrule
      
      \rowcolor{gray!20}
      \multicolumn{3}{c}{\textsc{Qwen3-8B}} \\
      \textsc{Qwen3-8B}       & 26.45\phantom{ (-0.00)} & 20.67\phantom{ (-0.00)} \\
      \textit{w/ }\textsc{TTRL}       & 47.13\phantom{ (-0.00)} & 27.40\phantom{ (-0.00)} \\
      \rowcolor{lightblue!100}
      \textit{w/ }\textsc{SCOPE}                            & 52.70\phantom{ (-0.00)} & 31.00\phantom{ (-0.00)} \\
      - \textit{w/o Conf}                & 47.70 (-5.00) & 28.36 (-2.64) \\
      - \textit{w/o Subgroup}            & 47.91 (-4.79) & 26.92 (-4.08) \\
      \bottomrule
    \end{tabular}
  }
  \caption{Ablation study on step-wise confidence (\textit{w/o Conf}) and automatic subgroup partition (\textit{w/o Subgroup}).}
  \label{tab:ablation}
\end{table}

Table~\ref{tab:ablation} shows the results. The consistent performance degradation upon removing either component validates the indispensability of our dual-granularity design. Specifically, the removal of subgroup partitioning leads to a sharp 5.83\% decline for Qwen2.5-1.5B on AIME 2024, exposing the exploration bottleneck caused by monolithic supervision. Similarly, relying on naive majority voting yields a 5.00\% deficit for Qwen3-8B on the same benchmark, confirming that naive majority voting is an unreliable proxy for reasoning quality. These findings indicate that model self-improvement hinges on the joint calibration of consensus scope and reward density.

\section{Analyses}

\subsection{Impact of the Trade-off Parameter on Quality-Diversity Balance}
\label{subsec:lambda_sensitivity}

\begin{figure}[t!]
    \centering
    \includegraphics[width=1.0\linewidth]{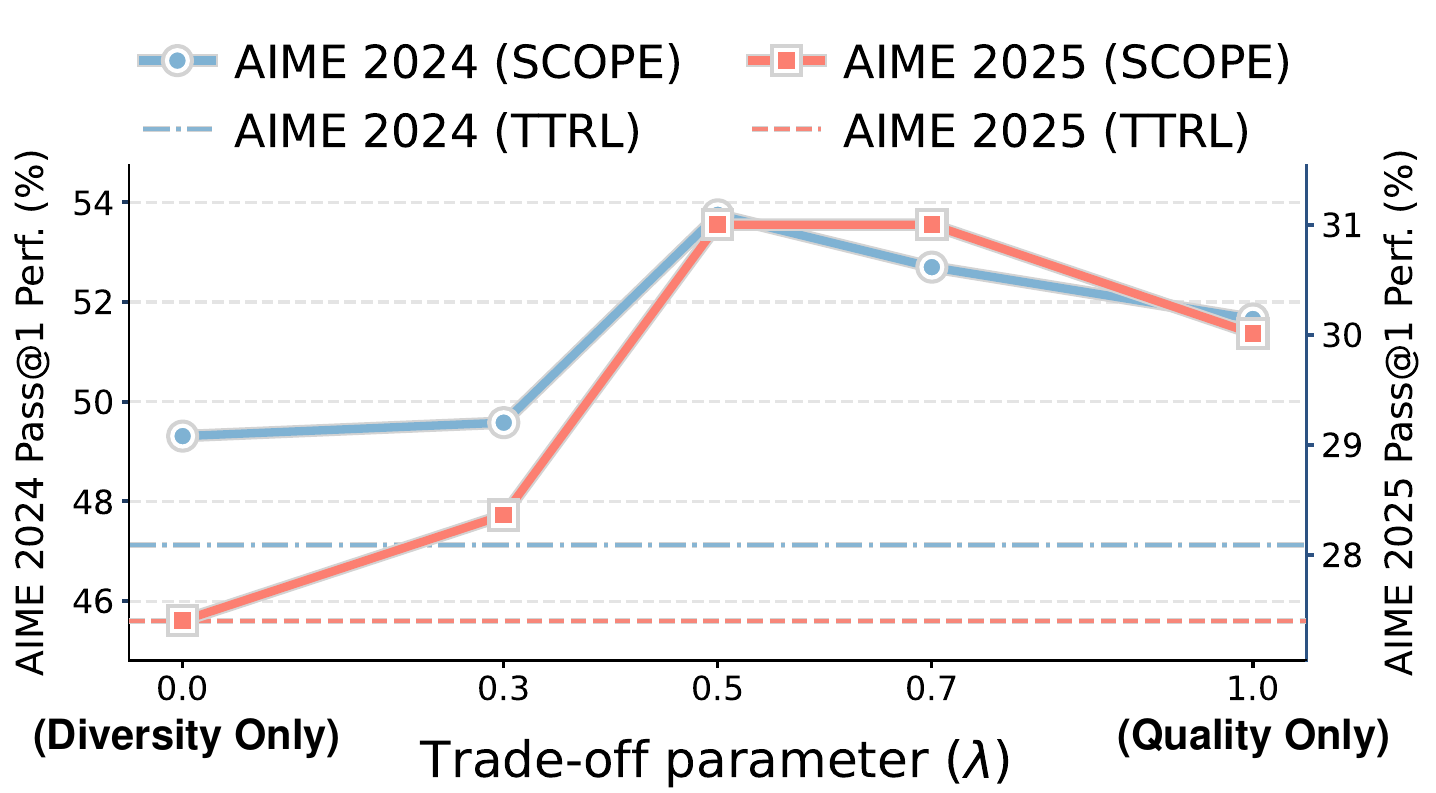}
    \caption{Analysis of the trade-off parameter \(\lambda\).}
    \label{fig:lambda_sensitivity}
\end{figure}

To investigate the trade-off between reasoning quality and diversity, we analyze the trade-off parameter \(\lambda\) in Eq.~\ref{eq:lambda} using Qwen3-8B. We vary \(\lambda\) from \(0.0\) to \(1.0\) and evaluate performance on AIME 2024 and AIME 2025. The results are shown in Figure~\ref{fig:lambda_sensitivity}.

When \(\lambda\) is set to \(1.0\), \textsc{SCOPE} relies solely on consensus quality. Although this setting already outperforms the TTRL baseline (horizontal dashed lines), performance saturates at 51.66\% on AIME 2024. Introducing exploration by decreasing \( \lambda \) yields consistent gains, with peak performance at \(\lambda=0.5\),  reaching 53.75\% on AIME 2024 and 31.0\% on AIME 2025. This indicates that while consensus is essential, encouraging exploration enables additional reasoning improvements beyond pure consensus guidance.

Conversely, setting \( \lambda \) too low (e.g., \( \lambda = 0 \)) results in a noticeable performance decline. Lacking the guidance of consensus quality, the optimization process is prone to over-exploration where the model drifts away from correct reasoning trajectories. This underscores the importance of jointly optimizing for both consensus alignment and exploration to achieve robust performance.

\subsection{Impact of Confidence Granularity}
\label{subsec:conf_granularity}

\begin{figure}[t!]
    \centering
    \includegraphics[width=0.9\linewidth]{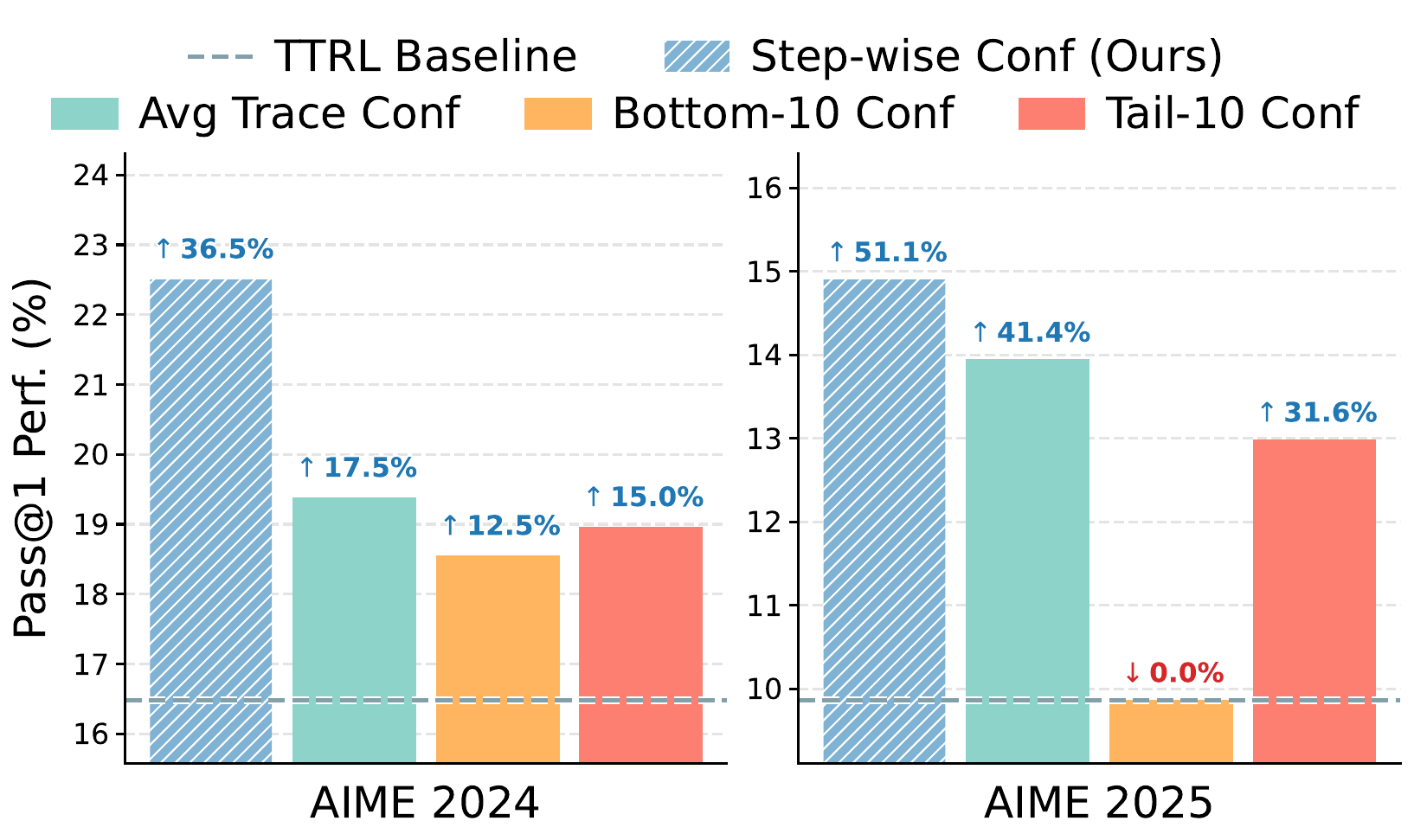}
    \caption{Impact analysis of confidence granularity. Comparison between \textsc{SCOPE} with our proposed step-wise confidence and alternative aggregation strategies. }
    \label{fig:conf_granularity}
\end{figure}

Figure~\ref{fig:conf_granularity} compares our step-wise confidence strategy with average trace confidence, bottom-10\% confidence, and tail-10\% confidence~\citep{fu2025deepParadigm} using Qwen2.5-Math-1.5B. Our method consistently outperforms all alternatives, achieving a 36.5\% relative improvement on AIME 2024 and a larger 51.1\% gain on the more challenging AIME 2025. In contrast, the bottom-10\% strategy collapses (+0.0\%), as sparse supervision overemphasizes the weakest step, penalizing difficult yet correct reasoning and discarding informative signals from the remainder of the chain.

Although the average trace strategy is more stable, it remains suboptimal due to error masking, where numerous trivial high-confidence steps obscure a critical intermediate mistake. Step-wise confidence overcomes this by enforcing dense, temporal supervision. By aligning reward resolution with step-level reasoning, our method ensures precise credit assignment, identifying and rectifying logical fallacies exactly where they occur without being smoothed out by global aggregation.

\subsection{Efficacy of Automatic Subgroup Selection}
\label{subsec:subgroup_selection}

\begin{figure}[t!]
    \centering
    \includegraphics[width=0.9\linewidth]{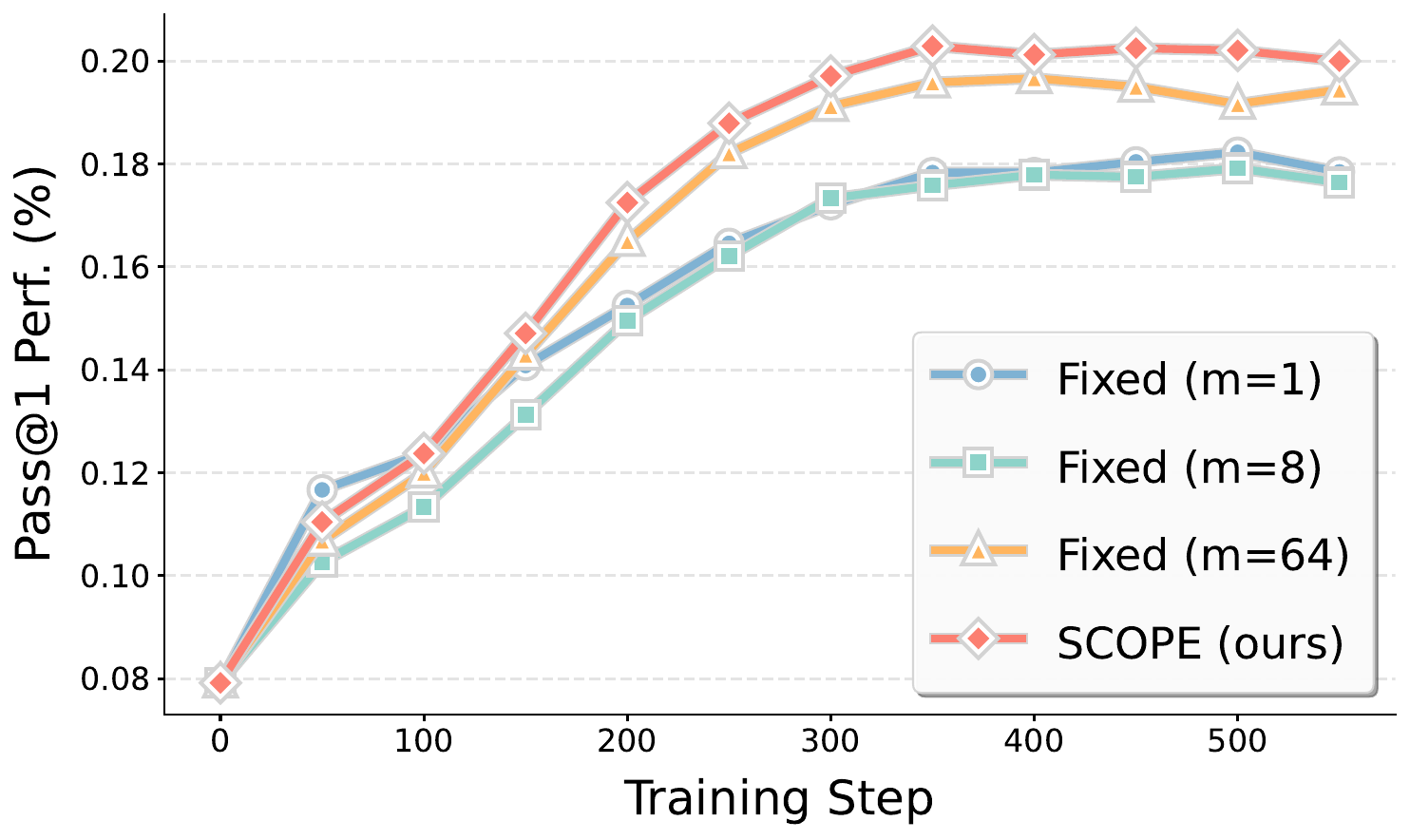}
    \caption{Impact analysis of subgroup size on training dynamics. Comparison between \textsc{SCOPE} with automatic subgroup size selection and variants with fixed subgroup sizes.}
    \label{fig:subgroup_selection}
\end{figure}

We analyze the training dynamics of \textsc{SCOPE} and fixed subgroup partition strategies on Qwen2.5-Math-1.5B across AIME 2024 and AIME 2025 to investigate the effectiveness of automatic subgroup size selection. Figure~\ref{fig:subgroup_selection} reveals a clear trade-off in consensus granularity. The atomized (\(m=1\)) and small subgroup (\(m=8\)) settings exhibit rapid initial growth but suffer from early saturation at a suboptimal level. This suggests that insufficient consensus amplifies noise and induces unreliable rewards and confirmation bias.

Conversely, while the global setting ($ m=64 $) ensures stability, it converges more slowly due to an exploration bottleneck induced by static subgroup partition. In contrast, the automatic selection strategy consistently leads, achieving both faster convergence and higher peak performance. By dynamically balancing exploration and quality, it avoids the noise of small groups while mitigating the rigidity of global consensus, thereby maximizing sample efficiency.

\subsection{Subgroup Partition Reduces Consensus Drift}
\label{subsec:consensus_drift}

To investigate whether subgroup partitioning reduces consensus drift compared with global consensus without subgrouping, we further analyze pseudo-label stability using Qwen2.5-Math-1.5B. We introduce \emph{Pseudo-label Accuracy} (PLA), which measures the strict consistency between estimated consensus pseudo-labels and ground-truth labels during training. Higher PLA indicates lower pseudo-label noise and a more reliable consensus signal. We compare TTRL with fixed subgroup sizes and the automatic partition version of \textsc{SCOPE}, with results reported in Table~\ref{tab:pla_consensus_drift}.

The results show that \textsc{SCOPE} consistently improves PLA over TTRL across different subgroup settings, demonstrating that subgroup partitioning reduces consensus drift rather than amplifying pseudo-label noise. Notably, the automatic partition version achieves the highest PLA, outperforming TTRL by 20.81\%. These results indicate that subgroup partitioning improves pseudo-label stability and effectively mitigates consensus drift.

\begin{table}[t!]
    \centering
    \small
    \begin{tabular}{lcc}
        \toprule
        \textbf{Method} & \textbf{PLA} & \textbf{Improvement} \\
        \midrule
        TTRL & 25.42 & -- \\
        \textsc{SCOPE} ($m=1$) & 26.75 & 5.24\% \\
        \textsc{SCOPE} ($m=8$) & 26.84 & 5.59\% \\
        \textsc{SCOPE} ($m=64$) & 29.57 & 16.33\% \\
        \textsc{SCOPE} (ours) & 30.71 & 20.81\% \\
        \bottomrule
    \end{tabular}
    \caption{Analysis of pseudo-label accuracy under different subgroup sizes.}
    \label{tab:pla_consensus_drift}
\end{table}

\section{Related Work}

\paragraph{RL for LLMs Reasoning.} 
Reinforcement learning has emerged as a critical paradigm for enhancing the reasoning capabilities of LLMs~\citep{zhang2023cumulative,guo2025deepseek}. Recent advancements focus on optimizing learning signals and training strategies. From the perspective of action space granularity, recent works have also explored extending single-token optimization to multi-token blocks to capture structural semantics~\citep{xu2026beyond}. To improve sample efficiency, \citet{wang2025reinforcement} introduced 1-shot RLVR, which selects high-quality examples based on historical variance to match the efficacy of large-scale training. For finer-grained credit assignment, \citet{wang2025beyond} leveraged chain-of-thought entropy to identify critical tokens for targeted policy updates. In terms of data curation, \citet{ye2025beyond} proposed the Process Consistency Filter to harmonize noisy signals by filtering samples based on process–outcome consistency. Furthermore, employing adversarial strategies, \citet{wu2025rlac} introduced RLAC, where a critic generates verifiable rubrics to guide the generator's optimization. Despite these significant strides, most of these methods still rely on substantial amounts of labeled data or ground-truth feedback during training.

\paragraph{Unsupervised RL.}
Unsupervised RL, which leverages self-derived signals without external annotation, has emerged as a promising direction for autonomous evolution~\citep{sun-etal-2025-self,ji-etal-2025-unlocking,chuang2025selfcite}. Early explorations validated this paradigm through different methods~\citep{wang2025improving,yuan2024selfrewarding}. Following the success of DeepSeek-R1~\citep{guo2025deepseek}, recent efforts have increasingly focused on circumventing the reliance on labeled data. For instance, \citet{zuo2025ttrl} introduced TTRL, employing majority-voted labels as reward proxies—a mechanism further refined by EVOL-RL~\citep{zhou2025evolving} via novelty incentives and Co-rewarding~\cite{zhang2025co} through semantic consistency checks. Alternatively, other works exploit internal model states: \citet{zhao2025learning} formulates self-certainty as an intrinsic reward for advantage estimation, while \citet{van2025post} utilize raw confidence scores as intrinsic feedback for preference optimization. However, relying solely on coarse-grained consensus or uncalibrated confidence signals often introduces noise and confirmation bias, failing to provide the fine-grained, reliable guidance necessary for solving complex reasoning tasks.

\section{Conclusion}

We propose \textsc{SCOPE}, a test-time reinforcement learning framework that mitigates confirmation bias and reward sparsity during unsupervised RL. By leveraging step-wise confidence and dynamic subgroup partitioning, \textsc{SCOPE} provides more reliable supervision and enables diverse, high-quality reasoning exploration. It achieves superior performance across representative reasoning benchmarks, consistently surpassing strong baselines.

\section*{Limitations}

Despite significant performance gains, \textsc{SCOPE} has certain limitations. First, our step definition relies on heuristic segmentation based on newline characters. While this aligns with the standard output format of most reasoning models, it assumes a structured generation pattern. Second, the dynamic calculation of Pareto-optimal subgroups introduces roughly a 10\% computational overhead. However, considering the substantial improvements in sample efficiency and final accuracy, we regard this as a highly favorable trade-off. Future work may extend this paradigm to longer-horizon and interactive agent settings, and explore more efficient test-time training dynamics to support robust self-evolution at scale.

\section*{Acknowledgments}

This work was supported in parts by ICFCRT (W2441020), Guangdong Basic and Applied Basic Research Foundation (2026A1515011358), Shenzhen Natural Science Foundation (JCYJ2025 0604181610014), and Scientific Development Funds from Shenzhen University.


\bibliography{custom}

@article{yang2024qwen2,
  title={Qwen2.5-math technical report: Toward mathematical expert model via self-improvement},
  author={Yang, An and Zhang, Beichen and Hui, Binyuan and Gao, Bofei and Yu, Bowen and Li, Chengpeng and Liu, Dayiheng and Tu, Jianhong and Zhou, Jingren and Lin, Junyang and others},
  journal={arXiv preprint arXiv:2409.12122},
  url={https://arxiv.org/abs/2409.12122},
  year={2024}
}

@article{grattafiori2024llama,
  title={The llama 3 herd of models},
  author={Grattafiori, Aaron and Dubey, Abhimanyu and Jauhri, Abhinav and Pandey, Abhinav and Kadian, Abhishek and Al-Dahle, Ahmad and Letman, Aiesha and Mathur, Akhil and Schelten, Alan and Vaughan, Alex and others},
  journal={arXiv preprint arXiv:2407.21783},
  url={https://arxiv.org/abs/2407.21783},
  year={2024}
}

@article{zhang2025co,
  title={Co-rewarding: Stable Self-supervised RL for Eliciting Reasoning in Large Language Models},
  author={Zhang, Zizhuo and Zhu, Jianing and Ge, Xinmu and Zhao, Zihua and Zhou, Zhanke and Li, Xuan and Feng, Xiao and Yao, Jiangchao and Han, Bo},
  journal={arXiv preprint arXiv:2508.00410},
  url={https://arxiv.org/abs/2508.00410},
  year={2025}
}

@article{zhou2025evolving,
  title={Evolving language models without labels: Majority drives selection, novelty promotes variation},
  author={Zhou, Yujun and Liang, Zhenwen and Liu, Haolin and Yu, Wenhao and Panaganti, Kishan and Song, Linfeng and Yu, Dian and Zhang, Xiangliang and Mi, Haitao and Yu, Dong},
  journal={arXiv preprint arXiv:2509.15194},
  url={https://arxiv.org/abs/2509.15194},
  year={2025}
}

@article{zhao2025learning,
  title={Learning to Reason without External Rewards},
  author={Zhao, Xuandong and Kang, Zhewei and Feng, Aosong and Levine, Sergey and Song, Dawn},
  journal={arXiv preprint arXiv:2505.19590},
  url={https://arxiv.org/abs/2505.19590},
  year={2025}
}

@article{wen2025reinforcement,
  title={Reinforcement Learning with Verifiable Rewards Implicitly Incentivizes Correct Reasoning in Base LLMs},
  author={Wen, Xumeng and Liu, Zihan and Zheng, Shun and Xu, Zhijian and Ye, Shengyu and Wu, Zhirong and Liang, Xiao and Wang, Yang and Li, Junjie and Miao, Ziming and others},
  journal={arXiv preprint arXiv:2506.14245},
  url={https://arxiv.org/abs/2506.14245},
  year={2025}
}

@article{su2025crossing,
  title={Crossing the Reward Bridge: Expanding RL with Verifiable Rewards Across Diverse Domains},
  author={Su, Yi and Yu, Dian and Song, Linfeng and Li, Juntao and Mi, Haitao and Tu, Zhaopeng and Zhang, Min and Yu, Dong},
  journal={arXiv preprint arXiv:2503.23829},
  url={https://arxiv.org/abs/2503.23829},
  year={2025}
}

@article{tang2025towards,
  title={Towards high data efficiency in reinforcement learning with verifiable reward},
  author={Tang, Xinyu and Zhang, Zhenduo and Liu, Yurou and Zhao, Wayne Xin and Wen, Zujie and Zhang, Zhiqiang and Zhou, Jun},
  journal={arXiv preprint arXiv:2509.01321},
  url={https://arxiv.org/abs/2509.01321},
  year={2025}
}

@article{fu2025deepParadigm,
  title={Deep think with confidence},
  author={Fu, Yichao and Wang, Xuewei and Tian, Yuandong and Zhao, Jiawei},
  journal={arXiv preprint arXiv:2508.15260},
  url={https://arxiv.org/abs/2508.15260},
  year={2025}
}

@article{guo2025deepseek,
  title={Deepseek-r1: Incentivizing reasoning capability in llms via reinforcement learning},
  author={DeepSeek-AI},
  journal={arXiv preprint arXiv:2501.12948},
  url={https://arxiv.org/abs/2501.12948},
  year={2025}
}

@article{yang2025qwen3,
  title={Qwen3 technical report},
  author={Yang, An and Li, Anfeng and Yang, Baosong and Zhang, Beichen and Hui, Binyuan and Zheng, Bo and Yu, Bowen and Gao, Chang and Huang, Chengen and Lv, Chenxu and others},
  journal={arXiv preprint arXiv:2505.09388},
  url={https://arxiv.org/abs/2505.09388},
  year={2025}
}

@article{van2025post,
  title={Post-Training Large Language Models via Reinforcement Learning from Self-Feedback},
  author={van Niekerk, Carel and Vukovic, Renato and Ruppik, Benjamin Matthias and Lin, Hsien-chin and Ga{\v{s}}i{\'c}, Milica},
  journal={arXiv preprint arXiv:2507.21931},
  url={https://arxiv.org/abs/2507.21931},
  year={2025}
}

@article{zhang2023cumulative,
  title={Cumulative Reasoning With Large Language Models},
  author={Zhang, Yifan and Yang, Jingqin and Yuan, Yang and Yao, Andrew Chi-Chih},
  journal={Transactions on Machine Learning Research; arXiv preprint arXiv:2308.04371},
  url={https://arxiv.org/abs/2308.04371},
  year={2023}
}

@article{yu2025rlpr,
  title={{RLPR}: Extrapolating RLVR to General Domains without Verifiers},
  author={Yu, Tianyu and Ji, Bo and Wang, Shouli and Yao, Shu and Wang, Zefan and Cui, Ganqu and Yuan, Lifan and Ding, Ning and Yao, Yuan and Liu, Zhiyuan and others},
  journal={arXiv preprint arXiv:2506.18254},
  url={https://arxiv.org/abs/2506.18254},
  year={2025}
}

@article{shao2024deepseekmath,
  title={Deepseekmath: Pushing the limits of mathematical reasoning in open language models},
  author={Shao, Zhihong and Wang, Peiyi and Zhu, Qihao and Xu, Runxin and Song, Junxiao and Bi, Xiao and Zhang, Haowei and Zhang, Mingchuan and Li, YK and Wu, Yang and others},
  journal={arXiv preprint arXiv:2402.03300},
  url={https://arxiv.org/abs/2402.03300},
  year={2024}
}

@article{li2024numinamath,
  title={Numinamath: The largest public dataset in ai4maths with 860k pairs of competition math problems and solutions},
  author={Li, Jia and Beeching, Edward and Tunstall, Lewis and Lipkin, Ben and Soletskyi, Roman and Huang, Shengyi and Rasul, Kashif and Yu, Longhui and Jiang, Albert Q and Shen, Ziju and others},
  journal={Hugging Face repository},
  url={https://huggingface.co/collections/AI-MO/numinamath},
  volume={13},
  number={9},
  pages={9},
  year={2024}
}

@article{chen2021evaluating,
  title={Evaluating large language models trained on code},
  author={Chen, Mark and Tworek, Jerry and Jun, Heewoo and Yuan, Qiming and Pinto, Henrique Ponde de Oliveira and Kaplan, Jared and Edwards, Harri and Burda, Yuri and Joseph, Nicholas and Brockman, Greg and others},
  journal={arXiv preprint arXiv:2107.03374},
  url={https://arxiv.org/abs/2107.03374},
  year={2021}
}

@article{wu2025rlac,
  title={{RLAC}: Reinforcement Learning with Adversarial Critic for Free-Form Generation Tasks},
  author={Wu, Mian and Zhang, Gavin and Min, Sewon and Levine, Sergey and Kumar, Aviral},
  journal={arXiv preprint arXiv:2511.01758},
  url={https://arxiv.org/abs/2511.01758},
  year={2025}
}

@article{lou2025adacot,
  title={{AdaCoT}: Pareto-Optimal Adaptive Chain-of-Thought Triggering via Reinforcement Learning},
  author={Lou, Chenwei and Sun, Zewei and Liang, Xinnian and Qu, Meng and Shen, Wei and Wang, Wenqi and Li, Yuntao and Yang, Qingping and Wu, Shuangzhi},
  journal={arXiv preprint arXiv:2505.11896},
  url={https://arxiv.org/abs/2505.11896},
  year={2025}
}

@article{ye2025beyond,
  title={Beyond Correctness: Harmonizing Process and Outcome Rewards through RL Training},
  author={Ye, Chenlu and Yu, Zhou and Zhang, Ziji and Chen, Hao and Sadagopan, Narayanan and Huang, Jing and Zhang, Tong and Beniwal, Anurag},
  journal={arXiv preprint arXiv:2509.03403},
  url={https://arxiv.org/abs/2509.03403},
  year={2025}
}

@article{jaech2024openai,
  title={Openai o1 system card},
  author={OpenAI},
  url={https://arxiv.org/abs/2412.16720},
  journal={arXiv preprint arXiv:2412.16720},
  year={2024}
}

@article{huang2026sat,
  title={SAT: Balancing Reasoning Accuracy and Efficiency with Stepwise Adaptive Thinking},
  author={Huang, Weiyang and Bai, Xuefeng and Chen, Kehai and Chen, Xinyang and Chen, Yibin and Guan, Weili and Zhang, Min},
  journal={arXiv preprint arXiv:2604.07922},
  url={https://arxiv.org/abs/2604.07922},
  year={2026}
}

@article{xu2026beyond,
  title={Beyond Token-Level Policy Gradients for Complex Reasoning with Large Language Models},
  author={Xu, Mufan and Chen, Kehai and Bai, Xuefeng and Niu, Zhengyu and Yang, Muyun and Zhao, Tiejun and Zhang, Min},
  journal={arXiv preprint arXiv:2602.14386},
  url={https://arxiv.org/abs/2602.14386},
  year={2026}
}

@inproceedings{sheng2024hybridflow,
  author={Guangming Sheng and Chi Zhang and Zilingfeng Ye and Xibin Wu and Wang Zhang and Ru Zhang and Yanghua Peng and Haibin Lin and Chuan Wu},
  title={HybridFlow: A Flexible and Efficient RLHF Framework},
  year={2025},
  cdate={1735689600000},
  pages={1279-1297},
  url={https://doi.org/10.1145/3689031.3696075},
  booktitle={EuroSys}
}

@inproceedings{NEURIPS2024_89f39d0b,
 author = {Zhong, Yifan and Ma, Chengdong and Zhang, Xiaoyuan and Yang, Ziran and Chen, Haojun and Zhang, Qingfu and Qi, Siyuan and Yang, Yaodong},
 booktitle = {Advances in Neural Information Processing Systems},
 doi = {10.52202/079017-2405},
 editor = {A. Globerson and L. Mackey and D. Belgrave and A. Fan and U. Paquet and J. Tomczak and C. Zhang},
 pages = {75522--75558},
 publisher = {Curran Associates, Inc.},
 title = {Panacea: Pareto Alignment via Preference Adaptation for LLMs},
 url = {https://proceedings.neurips.cc/paper_files/paper/2024/file/89f39d0b3d49a47606a165eefba2778c-Paper-Conference.pdf},
 volume = {37},
 year = {2024}
}

@inproceedings{
zeng2025simplerl,
title={Simple{RL}-Zoo: Investigating and Taming Zero Reinforcement Learning for Open Base Models in the Wild},
author={Weihao Zeng and Yuzhen Huang and Qian Liu and Wei Liu and Keqing He and Zejun MA and Junxian He},
booktitle={Second Conference on Language Modeling},
year={2025},
url={https://openreview.net/forum?id=vSMCBUgrQj}
}

@inproceedings{wang-etal-2025-ranked,
    title = "Ranked Voting based Self-Consistency of Large Language Models",
    author = "Wang, Weiqin  and
      Wang, Yile  and
      Huang, Hui",
    editor = "Che, Wanxiang  and
      Nabende, Joyce  and
      Shutova, Ekaterina  and
      Pilehvar, Mohammad Taher",
    booktitle = "Findings of the Association for Computational Linguistics: ACL 2025",
    month = jul,
    year = "2025",
    address = "Vienna, Austria",
    publisher = "Association for Computational Linguistics",
    url = "https://aclanthology.org/2025.findings-acl.744/",
    doi = "10.18653/v1/2025.findings-acl.744",
    pages = "14410--14426",
    ISBN = "979-8-89176-256-5"
}

@inproceedings{
chuang2025selfcite,
title={{SelfCite}: Self-Supervised Alignment for Context Attribution in Large Language Models},
author={Yung-Sung Chuang and Benjamin Cohen-Wang and Zejiang Shen and Zhaofeng Wu and Hu Xu and Xi Victoria Lin and James R. Glass and Shang-Wen Li and Wen-tau Yih},
booktitle={Forty-second International Conference on Machine Learning},
year={2025},
url={https://openreview.net/forum?id=rKi8eyJBoB}
}

@inproceedings{hendrycks2021measuring,
 author = {Hendrycks, Dan and Burns, Collin and Kadavath, Saurav and Arora, Akul and Basart, Steven and Tang, Eric and Song, Dawn and Steinhardt, Jacob},
 booktitle = {Proceedings of the Neural Information Processing Systems Track on Datasets and Benchmarks},
 editor = {J. Vanschoren and S. Yeung},
 pages = {},
 title = {Measuring Mathematical Problem Solving With the MATH Dataset},
 url = {https://datasets-benchmarks-proceedings.neurips.cc/paper_files/paper/2021/file/be83ab3ecd0db773eb2dc1b0a17836a1-Paper-round2.pdf},
 volume = {1},
 year = {2021}
}

@inproceedings{
wang2025beyond,
title={Beyond the 80/20 Rule: High-Entropy Minority Tokens Drive Effective Reinforcement Learning for {LLM} Reasoning},
author={Shenzhi Wang and Le Yu and Chang Gao and Chujie Zheng and Shixuan Liu and Rui Lu and Kai Dang and Xiong-Hui Chen and Jianxin Yang and Zhenru Zhang and Yuqiong Liu and An Yang and Andrew Zhao and Yang Yue and Shiji Song and Bowen Yu and Gao Huang and Junyang Lin},
booktitle={The Thirty-ninth Annual Conference on Neural Information Processing Systems},
year={2025},
url={https://openreview.net/forum?id=yfcpdY4gMP}
}

@inproceedings{ji-etal-2025-unlocking,
    title = "Unlocking {LLM}s' Self-Improvement Capacity with Autonomous Learning for Domain Adaptation",
    author = "Ji, Ke  and
      Chen, Junying  and
      Gao, Anningzhe  and
      Xie, Wenya  and
      Wan, Xiang  and
      Wang, Benyou",
    editor = "Che, Wanxiang  and
      Nabende, Joyce  and
      Shutova, Ekaterina  and
      Pilehvar, Mohammad Taher",
    booktitle = "Findings of the Association for Computational Linguistics: ACL 2025",
    month = jul,
    year = "2025",
    address = "Vienna, Austria",
    publisher = "Association for Computational Linguistics",
    url = "https://aclanthology.org/2025.findings-acl.1084/",
    doi = "10.18653/v1/2025.findings-acl.1084",
    pages = "21051--21067",
    ISBN = "979-8-89176-256-5"
}

@inproceedings{
zuo2025ttrl,
title={{TTRL}: Test-Time Reinforcement Learning},
author={Yuxin Zuo and Kaiyan Zhang and Li Sheng and Shang Qu and Ganqu Cui and Xuekai Zhu and Haozhan Li and Yuchen Zhang and Xinwei Long and Ermo Hua and Biqing Qi and Youbang Sun and Zhiyuan Ma and Lifan Yuan and Ning Ding and Bowen Zhou},
booktitle={The Thirty-ninth Annual Conference on Neural Information Processing Systems},
year={2025},
url={https://openreview.net/forum?id=VuVhgEiu20}
}

@inproceedings{pseudoLabel2019,
  title = {Pseudo-Labeling and Confirmation Bias in Deep Semi-Supervised Learning},
  author = {Eric Arazo and Diego Ortego and Paul Albert and Noel E O'Connor and Kevin McGuinness},
  booktitle={2020 International Joint Conference on Neural Networks (IJCNN)},
  url={https://ieeexplore.ieee.org/abstract/document/9207304},
  year={2020},
  organization={IEEE}
}

@inproceedings{
lightman2024lets,
title={Let's Verify Step by Step},
author={Hunter Lightman and Vineet Kosaraju and Yuri Burda and Harrison Edwards and Bowen Baker and Teddy Lee and Jan Leike and John Schulman and Ilya Sutskever and Karl Cobbe},
booktitle={The Twelfth International Conference on Learning Representations},
year={2024},
url={https://openreview.net/forum?id=v8L0pN6EOi}
}

@InProceedings{Prabhu_2021_ICCV,
    author    = {Prabhu, Viraj and Khare, Shivam and Kartik, Deeksha and Hoffman, Judy},
    title     = {SENTRY: Selective Entropy Optimization via Committee Consistency for Unsupervised Domain Adaptation},
    booktitle = {Proceedings of the IEEE/CVF International Conference on Computer Vision (ICCV)},
    url={https://openaccess.thecvf.com/content/ICCV2021/html/Prabhu_SENTRY_Selective_Entropy_Optimization_via_Committee_Consistency_for_Unsupervised_Domain_ICCV_2021_paper.html},
    month     = {October},
    year      = {2021},
    pages     = {8558-8567}
}

@inproceedings{sun-etal-2025-self,
    title = "The Self-Improvement Paradox: Can Language Models Bootstrap Reasoning Capabilities without External Scaffolding?",
    author = "Sun, Yutao  and
      Chen, Mingshuai  and
      Zhao, Tiancheng  and
      Xu, Ruochen  and
      Zhang, Zilun  and
      Yin, Jianwei",
    editor = "Che, Wanxiang  and
      Nabende, Joyce  and
      Shutova, Ekaterina  and
      Pilehvar, Mohammad Taher",
    booktitle = "Findings of the Association for Computational Linguistics: ACL 2025",
    month = jul,
    year = "2025",
    address = "Vienna, Austria",
    publisher = "Association for Computational Linguistics",
    url = "https://aclanthology.org/2025.findings-acl.337/",
    doi = "10.18653/v1/2025.findings-acl.337",
    pages = "6501--6512",
    ISBN = "979-8-89176-256-5"
}

@inproceedings{wang2025improving,
    title={Improving Rationality in the Reasoning Process of Language Models through Self-playing Game},
    author={Pinzheng Wang and Juntao Li and Zecheng Tang and Haijia Gui and Min zhang},
    booktitle={Forty-second International Conference on Machine Learning},
    year={2025},
    url={https://openreview.net/forum?id=PPsiS5nSlv}
}

@book{pareto1964cours,
  title={Cours d'{\'e}conomie politique},
  author={Pareto, Vilfredo},
  volume={1},
  url={https://shs.cairn.info/cours-d-economie-politique-tomes-1-et-2--9782600040143?tab=sommaire},
  year={1964},
  publisher={Librairie Droz}
}

@inproceedings{
wang2025reinforcement,
title={Reinforcement Learning for Reasoning in Large Language Models with One Training Example},
author={Yiping Wang and Qing Yang and Zhiyuan Zeng and Liliang Ren and Liyuan Liu and Baolin Peng and Hao Cheng and Xuehai He and Kuan Wang and Jianfeng Gao and Weizhu Chen and Shuohang Wang and Simon Shaolei Du and yelong shen},
booktitle={The Thirty-ninth Annual Conference on Neural Information Processing Systems},
year={2025},
url={https://openreview.net/forum?id=IBrRNLr6JA}
}

@inproceedings{
yuan2024selfrewarding,
title={Self-Rewarding Language Models},
author={Weizhe Yuan and Richard Yuanzhe Pang and Kyunghyun Cho and Xian Li and Sainbayar Sukhbaatar and Jing Xu and Jason E Weston},
booktitle={Forty-first International Conference on Machine Learning},
year={2024},
url={https://openreview.net/forum?id=0NphYCmgua}
}

\appendix

\section{Checkpoints of Models}
\label{app:model_links}
The checkpoints of open-source models in our experiments are shown in Table~\ref{tab:model_links}.

\begin{table}[h]
  \centering
  \small
  \resizebox{\linewidth}{!}{
    \begin{tabular}{ll}
      \toprule
      \textbf{Model} & \textbf{Resource Link} \\
      
      \midrule
      \rowcolor{gray!20}
      \multicolumn{2}{c}{\textsc{Qwen Series}} \\
      \textsc{Qwen2.5-Math-1.5B} & \href{https://huggingface.co/Qwen/Qwen2.5-Math-1.5B}{Qwen/Qwen2.5-Math-1.5B} \\
      \textsc{Qwen3-1.7B}        & \href{https://huggingface.co/Qwen/Qwen3-1.7B}{Qwen/Qwen3-1.7B} \\
      \textsc{Qwen3-8B}          & \href{https://huggingface.co/Qwen/Qwen3-8B}{Qwen/Qwen3-8B} \\
      
      \midrule
      \rowcolor{gray!20}
      \multicolumn{2}{c}{\textsc{Llama Series}} \\
      \textsc{LLaMA3.1-8B-Instruct} & \href{https://huggingface.co/meta-llama/Llama-3.1-8B-Instruct}{meta-llama/Llama-3.1-8B-Instruct} \\
      \bottomrule
    \end{tabular}
  }
  \caption{Checkpoints of open-source models in our experiments. }
  \label{tab:model_links}
\end{table}

\section{Example of Datasets}
\label{app:data_examples}

Table~\ref{tab:data_example} provides representative examples from the evaluation datasets along with their respective sizes.

\begin{table}[h]
  \centering
  \large
  \renewcommand{\arraystretch}{1.4}
  
  \resizebox{\linewidth}{!}{
    \begin{tabular}{lp{10cm}}
      \toprule
      \textbf{Type} & \textbf{Content} \\
      
      \midrule
      
      \rowcolor{gray!20}
      \multicolumn{2}{c}{\textbf{AIME 2024 (30)}} \\ 
      \textbf{Question} & Every morning Aya goes for a $9$-kilometer-long walk and stops at a coffee shop afterwards. When she walks at a constant speed of $s$ kilometers per hour, the walk takes her 4 hours, including $t$ minutes spent in the coffee shop. When she walks $s+2$ kilometers per hour, the walk takes her 2 hours and 24 minutes, including $t$ minutes spent in the coffee shop. Suppose Aya walks at $s+\frac{1}{2}$ kilometers per hour. Find the number of minutes the walk takes her, including the $t$ minutes spent in the coffee shop. \\
      \textbf{Answer}   & \boxed{204} \\

      \midrule
      \rowcolor{gray!20}
      \multicolumn{2}{c}{\textbf{AIME 2025 (30)}} \\
      \textbf{Question} & Find the sum of all integer bases $b>9$ for which $17_b$ is a divisor of $97_b.$ \\
      \textbf{Answer}   & \boxed{70} \\

      \midrule
      \rowcolor{gray!20}
      \multicolumn{2}{c}{\textbf{AMC (83)}} \\
      \textbf{Question} & $\frac{m}{n}$ is the Irreducible fraction value of \[3+\frac{1}{3+\frac{1}{3+\frac{1}{3}}}\], what is the value of $m+n$? \\
      \textbf{Answer}   & \boxed{142.0} \\
      
      \midrule
      \rowcolor{gray!20}
      \multicolumn{2}{c}{\textbf{MATH-500 (500)}} \\
      \textbf{Question} & Convert the point $(0,3)$ in rectangular coordinates to polar coordinates.  Enter your answer in the form $(r,\theta),$ where $r > 0$ and $0 \le \theta < 2 \pi.$ \\
      \textbf{Answer}   & \boxed{\left( 3, \frac{\pi}{2} \right)} \\
      \bottomrule
    \end{tabular}
  }
  \caption{Examples of questions and answers from the evaluation datasets.}
  \label{tab:data_example}
\end{table}

\begin{tcolorbox}[
    enhanced, 
    attach boxed title to top center={yshift=-3mm}, 
    colback=boxback, 
    colframe=boxframe, 
    coltitle=white, 
    title=\textbf{System Prompt}, 
    fonttitle=\bfseries, 
    boxed title style={
        colback=boxframe, 
        sharp corners, 
        frame hidden, 
    },
    sharp corners=south, 
    arc=3mm, 
    drop shadow, 
    boxrule=0.5mm, 
    top=1.5em, 
]
    \large Please reason step by step, and put your final answer within \textbackslash boxed\{\}.
\end{tcolorbox}

\section{Implementation Details}
\label{app:exp}

\paragraph{Training Configuration} We implement our method using the \textit{Volcano Engine Reinforcement Learning for LLMs} framework~\citep{sheng2024hybridflow}. Regarding hyperparameters, we employ the AdamW optimizer for the policy model, utilizing a cosine learning rate schedule with a peak value of $5 \times 10^{-7}$. During the rollout phase, we sample 64 responses per prompt using a temperature of 0.6 (adjusted to 1.0 for Qwen2.5-Math). For label estimation, we perform bootstrap sampling with 32 samples for each subgroup and subsequently utilize all 64 responses for training. The maximum generation length is set to 3,072 tokens.

\paragraph{Prompt for Training}
For all experiments, we employed a standardized system prompt to regulate the output format, explicitly requiring the model to articulate a step-by-step reasoning process followed by a clearly delimited final answer~\citep{zeng2025simplerl}:

\paragraph{Evaluation Configuration}

We apply our method to each benchmark individually, setting the maximum generation length to 3072 tokens unless otherwise specified. For the main experiments, following the protocol of DeepSeek-R1 \citep{guo2025deepseek}, we adopt the pass@$k$ metric \citep{chen2021evaluating} and report pass@1 using non-zero temperature sampling. Specifically, we generate 16 responses (4 for models with 32k context) per question using a temperature of 0.6 and a top-$p$ value of 0.95. The pass@1 score is computed as:
\begin{equation}
    \text{pass@1} = \frac{1}{k} \sum_{i=1}^{k} p_i,
\end{equation}
where \( p_i \) indicates correctness of the i-th response.
where \( p_i \) indicates whether the \( i \)-th response is correct.

\section{Case Study}
\label{app:case_study}

\begin{figure}[ht]
    \centering
    \begin{tcolorbox}[
        enhanced, 
        attach boxed title to top center={yshift=-3mm}, 
        colback=boxback, 
        colframe=boxframe, 
        coltitle=white, 
        title=\textbf{Question}, 
        fonttitle=\bfseries, 
        boxed title style={
            colback=boxframe, 
            sharp corners, 
            frame hidden, 
        },
        sharp corners=south, 
        arc=3mm, 
        drop shadow, 
        boxrule=0.5mm, 
        top=1.5em, 
    ]
        Every morning Aya goes for a $9$-kilometer-long walk and stops at a coffee shop afterwards. When she walks at a constant speed of $s$ kilometers per hour, the walk takes her 4 hours, including $t$ minutes spent in the coffee shop. When she walks $s+2$ kilometers per hour, the walk takes her 2 hours and 24 minutes, including $t$ minutes spent in the coffee shop. Suppose Aya walks at $s+\frac{1}{2}$ kilometers per hour. Find the number of minutes the walk takes her, including the $t$ minutes spent in the coffee shop.
        \label{question}
    \end{tcolorbox}
    \caption{A question from AIME 2024.}
    \label{fig:question}
\end{figure}

We present the generation outputs for the same question shown in Figure~\ref{fig:question} from four different models: (1) the backbone model (Figure~\ref{fig:solution1}); (2) the model after \textsc{TTRL} training (Figure~\ref{fig:solution2}); (3) the model after full \textsc{SCOPE} training (Figure~\ref{fig:solution3}). These correspond to solution1, solution2, and solution3, respectively. The backbone model (solution1) fails due to a critical logic error: adding total time instead of the break duration and hallucinates a final answer of \texttt{120}. The model trained with \textsc{TTRL} (solution2), although producing the correct numerical answer, exhibits clear flaws in mathematical modeling. Specifically: (1) in the \textbf{equation-formulation stage}, it erroneously subtracts the coffee-break time from the walking time (e.g., $\frac{9 \times 60}{s} - t = 240$); (2) in the \textbf{intermediate-calculation stage}, this mistake leads to an impossible negative duration ($t = -24$ minutes); and (3) in the \textbf{Python-execution stage}, the error is obscured by applying \texttt{abs()}, which masks the incorrect reasoning rather than correcting it. These incorrect components are highlighted with \fcolorbox{red}{white}{red boxes} in the annotated solution. In contrast, \textsc{SCOPE} (Solution 3) demonstrates superior reasoning by identifying the invariance of break time. It eliminates variables to form a concise quadratic equation, grasping the true physical relationships without erroneous patching. It efficiently eliminates the variable $t$ to form a concise quadratic equation, showcasing that \textsc{SCOPE} enables the model to grasp the true physical relationships within the problem.

\section{Use of AI Assistants}
AI assistants were used solely for translation and linguistic polishing purposes. They did not contribute to the generation of research conclusions, experimental results, or the core technical content of this work.

\begin{figure*}[t!]
\centering
\begin{tcolorbox}[colback=boxback, colframe=boxframe, coltitle=white, title=\textbf{Solution1}]
\small
\setlength{\abovedisplayskip}{2pt plus 1pt minus 1pt}
\setlength{\belowdisplayskip}{2pt plus 1pt minus 1pt}
\setlength{\abovedisplayshortskip}{2pt plus 1pt}
\setlength{\belowdisplayshortskip}{2pt plus 1pt minus 0pt}
\setlength{\parskip}{1pt}
Let's solve the problem step by step using Python and SymPy.

\textbf{1. Understand the Problem:}
\begin{itemize}[noitemsep, topsep=0pt, leftmargin=*]
    \item When Aya walks at a speed of $s$ km/h, the total time taken for the walk and the coffee shop break is 4 hours.
    \item When Aya walks at a speed of $s+2$ km/h, the total time taken for the walk and the coffee shop break is 2 hours and 24 minutes (which is 2.4 hours).
    \item We need to find the time taken when Aya walks at a speed of $s + \frac{1}{2}$ km/h.
\end{itemize}

\textbf{2. Formulate the Equations:}
\begin{itemize}[noitemsep, topsep=0pt, leftmargin=*]
    \item Let $t$ be the time spent in the coffee shop break in minutes.
    \item When Aya walks at $s$ km/h, the time taken for the walk is $\frac{9}{s}$ hours, which is $\frac{9}{s} \times 60$ minutes. So the total time is $\frac{9}{s} \times 60 + t = 4 \times 60 = 240$ minutes.
    \item When Aya walks at $s+2$ km/h, the time taken for the walk is $\frac{9}{s+2}$ hours, which is $\frac{9}{s+2} \times 60$ minutes. So the total time is $\frac{9}{s+2} \times 60 + t = 2.4 \times 60 = 144$ minutes.
\end{itemize}

\textbf{3. Set Up the Equations:}
\begin{itemize}[noitemsep, topsep=0pt, leftmargin=*]
    \item From the first condition: $\frac{9}{s} \times 60 + t = 240$
    \item From the second condition: $\frac{9}{s+2} \times 60 + t = 144$
\end{itemize}

\textbf{4. Solve the Equations:}
\begin{itemize}[noitemsep, topsep=0pt, leftmargin=*]
    \item Subtract the second equation from the first to eliminate $t$:
    \[ \frac{9}{s} \times 60 + t - \left(\frac{9}{s+2} \times 60 + t\right) = 240 - 144 \]
    \item Simplify and solve for $s$:
    \begin{align*}
        \frac{9}{s} \times 60 - \frac{9}{s+2} \times 60 &= 96 \\
        9 \times 60 \left(\frac{1}{s} - \frac{1}{s+2}\right) &= 96 \\
        540 \left(\frac{s+2 - s}{s(s+2)}\right) &= 96 \implies \frac{1080}{s(s+2)} = 96 \\
        1080 &= 96s^2 + 192s \\
        96s^2 + 192s - 1080 &= 0
    \end{align*}
    \item Divide the whole equation by 24:
    \[ 4s^2 + 8s - 45 = 0 \]
    \item Solve this quadratic equation using the quadratic formula $s = \frac{-b \pm \sqrt{b^2 - 4ac}}{2a}$, where $a = 4$, $b = 8$, and $c = -45$.
\end{itemize}

\textbf{5. Find the Speed and Time:}
\begin{itemize}[noitemsep, topsep=0pt, leftmargin=*]
    \item Solve the quadratic equation to find $s$.
    \item Use the value of $s$ to find the time taken when walking at $s + \frac{1}{2}$ km/h.
\end{itemize}

Let's implement this in Python:
\begin{lstlisting}[language=Python, basicstyle=\ttfamily\scriptsize, keywordstyle=\color{blue}, frame=none, aboveskip=2pt, belowskip=0pt]
import sympy as sp
# Define the variable
s = sp.symbols('s')
# Define the quadratic equation
equation = 4*s**2 + 8*s - 45
# Solve the quadratic equation
solutions = sp.solve(equation, s)
# We need the positive solution
s_value = [sol for sol in solutions if sol > 0][0]
# Calculate the time taken when walking at s + 1/2 km/h
time_walk = (9 / (s_value + 0.5)) * 60
# The total time including the coffee shop break
total_time = time_walk + 144
# Convert the total time from minutes to hours and then to minutes
total_time_minutes = total_time
print(total_time_minutes)
\end{lstlisting}

\noindent\textbf{Output:}
\texttt{120}

The number of minutes the walk takes Aya, including the coffee shop break, when she walks at $s + \frac{1}{2}$ kilometers per hour is \boxed{120}.
\end{tcolorbox}
\caption{Solution from Qwen2.5-Math-1.5B.}
\label{fig:solution1}
\end{figure*}

\begin{figure*}[t!]
\centering
\begin{tcolorbox}[colback=boxback, colframe=boxframe, coltitle=white, title=\textbf{Solution2}]
\small
\setlength{\abovedisplayskip}{2pt plus 1pt minus 1pt}
\setlength{\belowdisplayskip}{2pt plus 1pt minus 1pt}
\setlength{\abovedisplayshortskip}{2pt plus 1pt}
\setlength{\belowdisplayshortskip}{2pt plus 1pt minus 0pt}
\setlength{\parskip}{1pt}
Let's break down the problem step-by-step and solve it using Python and SymPy.

\textbf{1. Understand the Problem:}
\begin{itemize}[noitemsep, topsep=0pt, leftmargin=*]
    \item When Aya walks at a speed of $s$ km/h, the walk takes her 4 hours, including $t$ minutes spent in the coffee shop.
    \item When she walks at a speed of $s+2$ km/h, the walk takes her 2 hours and 24 minutes, including $t$ minutes spent in the coffee shop.
    \item We need to find the number of minutes the walk takes her, including the $t$ minutes spent in the coffee shop, when she walks at $s+\frac{1}{2}$ km/h.
\end{itemize}

\textbf{2. Formulate the Equations:}
\begin{itemize}[noitemsep, topsep=0pt, leftmargin=*]
    \item The total time spent walking is $4$ hours when the speed is $s$ km/h.
    \item The total time spent walking is $2 + \frac{24}{60} = 2.4$ hours when the speed is $s+2$ km/h.
    \item The distance Aya walks is $9$ km in both cases.
    \item Let $t$ be the time spent in the coffee shop in minutes. Then the walking time is $4 \times 60 - t$ minutes when the speed is $s$ km/h, and $2.4 \times 60 - t$ minutes when the speed is $s+2$ km/h.
\end{itemize}

\textbf{3. Set Up the Equations:}
\begin{itemize}[noitemsep, topsep=0pt, leftmargin=*]
    \item When the speed is $s$ km/h, the walking time is $\frac{9}{s} \times 60 - t = 240 - t$ minutes.
    \item When the speed is $s+2$ km/h, the walking time is $\frac{9}{s+2} \times 60 - t = 144 - t$ minutes.
    \item So we have the following equations:
    \[ \frac{9 \times 60}{s} \fcolorbox{red}{white}{- t} = 240 \]
    \[ \frac{9 \times 60}{s+2} \fcolorbox{red}{white}{- t} = 144 \]
\end{itemize}

\textbf{4. Solve the Equations:}
\begin{itemize}[noitemsep, topsep=0pt, leftmargin=*]
    \item From the first equation, we get: $t = \frac{540}{s} - 240$.
    \item From the second equation, we get: $t = \frac{540}{s+2} - 144$.
    \item Equate the two expressions for $t$ and simplify:
    \begin{align*}
        \frac{540}{s} - 240 &= \frac{540}{s+2} - 144 \\
        540 \left( \frac{s+2 - s}{s(s+2)} \right) &= 96 \implies 1080 = 96s(s+2) \\
        4s^2 + 8s - 45 &= 0
    \end{align*}
    \item Solve the quadratic equation using the quadratic formula:
    \[ s = \frac{-8 \pm \sqrt{64 + 720}}{8} = \frac{20}{8} = 2.5 \text{ (since } s>0 \text{)} \]
\end{itemize}

\textbf{5. Find the Time Spent in the Coffee Shop:}
\begin{itemize}[noitemsep, topsep=0pt, leftmargin=*]
    \item Substitute $s = 2.5$ into the equation for $t$:
    \[ t = \frac{540}{2.5} - 240 = 216 - 240 = \fcolorbox{red}{white}{-24} \]
    \item $t = 24$ minutes (taking magnitude).
\end{itemize}

\textbf{6. Find the Time for the Walk at $s+\frac{1}{2}$ km/h:}
\begin{itemize}[noitemsep, topsep=0pt, leftmargin=*]
    \item The walking time is $\frac{9}{2.5 + 0.5} \times 60 = 180$ minutes.
    \item Including the $t$ minutes spent in the coffee shop, total time is $180 + 24 = 204$ minutes.
\end{itemize}

Let's implement this in Python to verify our solution.
\begin{lstlisting}[language=Python, basicstyle=\ttfamily\scriptsize, keywordstyle=\color{blue}, frame=none, aboveskip=2pt, belowskip=0pt, escapechar=|]
import sympy as sp
s = sp.symbols('s')
# Define equations based on derivation
eq1 = sp.Eq(540/s |\fcolorbox{red}{white}{-}| 240, 540/(s + 2) |\fcolorbox{red}{white}{-}| 144)    |\textcolor{red}{\textbf{<-- ERROR: flawed time equation}}|
solution = sp.solve(eq1, s)
s_val = [sol.evalf() for sol in solution if sol > 0][0]
t = |\fcolorbox{red}{white}{abs}|((540 / s_val) - 240)                     |\textcolor{red}{\textbf{<-- ERROR: masking negative time with abs()}}|
total = (9 / (s_val + 0.5)) * 60 + t
print(total)
\end{lstlisting}

\noindent\textbf{Output:}
\texttt{204.000000000000}

The total time Aya spends walking, including the time spent in the coffee shop, when she walks at a speed of $s + \frac{1}{2}$ km/h is \boxed{204} minutes.
\end{tcolorbox}
\caption{Solution after training process of \textsc{TTRL}.}
\label{fig:solution2}
\end{figure*}

\begin{figure*}[t!]
\centering
\begin{tcolorbox}[colback=boxback, colframe=boxframe, coltitle=white, title=\textbf{Solution3}]
\small
\setlength{\abovedisplayskip}{2pt plus 1pt minus 1pt}
\setlength{\belowdisplayskip}{2pt plus 1pt minus 1pt}
\setlength{\abovedisplayshortskip}{2pt plus 1pt}
\setlength{\belowdisplayshortskip}{2pt plus 1pt minus 0pt}
\setlength{\parskip}{1pt}
To solve this problem, we need to determine the time Aya spends walking and the time she spends in the coffee shop. Let's break it down step-by-step.

\textbf{1. Determine the time spent in the coffee shop:}
\begin{itemize}[noitemsep, topsep=0pt, leftmargin=*]
    \item When Aya walks at a speed of $s$ km/h, the total time for the walk including the coffee shop break is 4 hours. Since the walk itself takes $\frac{9}{s}$ hours, the time spent in the coffee shop is $4 - \frac{9}{s}$ hours.
    \item When Aya walks at a speed of $s+2$ km/h, the total time for the walk including the coffee shop break is 2 hours and 24 minutes, which is $2.4$ hours. The time spent in the coffee shop is $2.4 - \frac{9}{s+2}$ hours.
\end{itemize}
Since the time spent in the coffee shop is the same in both cases, we can set up the following equation:
\[ 4 - \frac{9}{s} = 2.4 - \frac{9}{s+2} \]

\textbf{2. Solve for $s$:}
Let's solve this equation step-by-step.
\begin{align*}
    4 - \frac{9}{s} &= 2.4 - \frac{9}{s+2} \\
    4 - 2.4 &= \frac{9}{s} - \frac{9}{s+2} \\
    1.6 &= 9 \left( \frac{1}{s} - \frac{1}{s+2} \right) \\
    1.6 &= 9 \left( \frac{(s+2) - s}{s(s+2)} \right) \\
    1.6 &= 9 \left( \frac{2}{s(s+2)} \right) \\
    1.6 &= \frac{18}{s(s+2)} \\
    1.6s(s+2) &= 18 \\
    1.6s^2 + 3.2s - 18 &= 0 \\
    16s^2 + 32s - 180 &= 0 \\
    4s^2 + 8s - 45 &= 0
\end{align*}
We can solve this quadratic equation using the quadratic formula $s = \frac{-b \pm \sqrt{b^2 - 4ac}}{2a}$, where $a = 4$, $b = 8$, and $c = -45$.
\begin{align*}
    s &= \frac{-8 \pm \sqrt{8^2 - 4 \cdot 4 \cdot (-45)}}{2 \cdot 4} \\
    s &= \frac{-8 \pm \sqrt{64 + 720}}{8} \\
    s &= \frac{-8 \pm \sqrt{784}}{8} = \frac{-8 \pm 28}{8}
\end{align*}
\[ s = \frac{20}{8} = 2.5 \quad \text{or} \quad s = \frac{-36}{8} = -4.5 \]
Since $s$ represents a speed, it must be positive. Therefore, $s = 2.5$ km/h.

\textbf{3. Determine the time spent in the coffee shop:}
\begin{itemize}[noitemsep, topsep=0pt, leftmargin=*]
    \item When Aya walks at $s = 2.5$ km/h, the time spent in the coffee shop is $4 - \frac{9}{2.5} = 4 - 3.6 = 0.4$ hours, which is $0.4 \times 60 = 24$ minutes.
\end{itemize}

\textbf{4. Determine the time for the walk at $s + \frac{1}{2} = 3$ km/h:}
\begin{itemize}[noitemsep, topsep=0pt, leftmargin=*]
    \item The time for the walk is $\frac{9}{3} = 3$ hours.
    \item Including the 24 minutes spent in the coffee shop, the total time is $3 + 0.4 = 3.4$ hours, which is $3.4 \times 60 = 204$ minutes.
\end{itemize}

The final answer is \boxed{204}.
\end{tcolorbox}
\caption{Solution after training process of \textsc{SCOPE}.}
\label{fig:solution3}
\end{figure*}

\end{document}